\documentclass{article} 
\usepackage{iclr2023_conference,times}

\usepackage{amsmath,amsfonts,bm}

\def\eqref#1{equation~\ref{#1}}

\def\1{\bm{1}}

\DeclareMathAlphabet{\mathsfit}{\encodingdefault}{\sfdefault}{m}{sl}
\SetMathAlphabet{\mathsfit}{bold}{\encodingdefault}{\sfdefault}{bx}{n}

\usepackage{hyperref}
\usepackage{url}

\usepackage{booktabs}
\usepackage{xspace}
\usepackage{multirow, makecell}
\usepackage{color, colortbl}
\usepackage{graphicx}
\usepackage{wrapfig}
\usepackage{mathtools}
\usepackage{caption}
\usepackage{subcaption}
\hypersetup{
    colorlinks=true,
    citecolor=teal,
    urlcolor=blue,
    }

\usepackage{pifont}
\newcommand{\xmark}{\ding{55}}%
\newcommand{\cmark}{\ding{51}}%

\newcommand{\OURS}{F-VLM}
\newcommand{\x}{{$\times$}}

\makeatletter
\DeclareRobustCommand\onedot{\futurelet\@let@token\@onedot}
\def\@onedot{\ifx\@let@token.\else.\null\fi\xspace}

\def\eg{\emph{e.g}\onedot} 
\def\ie{\emph{i.e}\onedot} 
 
\def\etc{\emph{etc}\onedot}

\def\etal{\emph{et al}\onedot}

\definecolor{baselinecolor}{gray}{.9}
\newcommand{\baseline}[1]{\cellcolor{baselinecolor}{#1}}

\makeatletter\renewcommand\paragraph{\@startsection{paragraph}{4}{\z@}
  {.0em \@plus.0ex \@minus.2ex}{-.5em}{\normalfont\normalsize\bfseries}}\makeatother

\title{F-VLM: Open-Vocabulary Object Detection upon Frozen Vision and Language Models}

\author{
Weicheng Kuo$^\star$, Yin Cui$^\dagger$, Xiuye Gu$^\dagger$, AJ Piergiovanni$^\star$, Anelia Angelova$^\star$\\
$^\star$Google Research, Brain Team; $^\dagger$Google Research, Perception\\
\texttt{\{weicheng, yincui, xiuyegu, ajpiergi, anelia\}@google.com}
}

\iclrfinalcopy 
\begin{document}

\maketitle

\begin{abstract}
We present \textbf{\OURS{}}, a simple open-vocabulary object detection method built upon \textbf{F}rozen \textbf{V}ision and \textbf{L}anguage \textbf{M}odels. \OURS{} simplifies the current multi-stage training pipeline by eliminating the need for knowledge distillation or detection-tailored pretraining. Surprisingly, we observe that a \textit{frozen} VLM: 1) retains the locality-sensitive features necessary for detection, and 2) is a strong region classifier. We finetune only the detector head and combine the detector and VLM outputs for each region at inference time. 
\OURS{} shows compelling scaling behavior and achieves +6.5 mask AP improvement over the previous state-of-the-art on LVIS open-vocabulary detection benchmark at system level. In addition, we demonstrate very competitive results on COCO open-vocabulary detection benchmark and cross-dataset transfer detection, in addition to significant training speed-up and compute savings. The code will be released~\footnote{Project page:~\href{https://sites.google.com/view/f-vlm/home}{https://sites.google.com/view/f-vlm/home}}.
\end{abstract}

\vspace{-4mm}
\section{Introduction}
\vspace{-2mm}

Detection is a fundamental vision task that aims to localize and recognize objects in an image. 
However, the data collection process of manually annotating bounding boxes or instance masks is tedious and costly, which limits the modern detection vocabulary size to an order of $10^3$. This is orders of magnitude smaller than the vocabulary humans use to describe the visual world.
To overcome such limitation, we focus on open-vocabulary object detection~\citep{Zareian_2021_CVPR,gu2022openvocabulary} to take detection beyond a fixed set of vocabulary.

Recently, vision and language models (VLMs) have gained strong open-vocabulary visual recognition capability by learning from Internet-scale image-text pairs~\citep{radford2021clip,align}. They are typically applied to zero-shot classification (\eg, on ImageNet) using frozen weights without finetuning, which stands in stark contrast to the existing paradigms of retraining or finetuning when applying VLMs for open-vocabulary detection. 

\begin{figure*}[t]
    \includegraphics[width=1.00\linewidth]{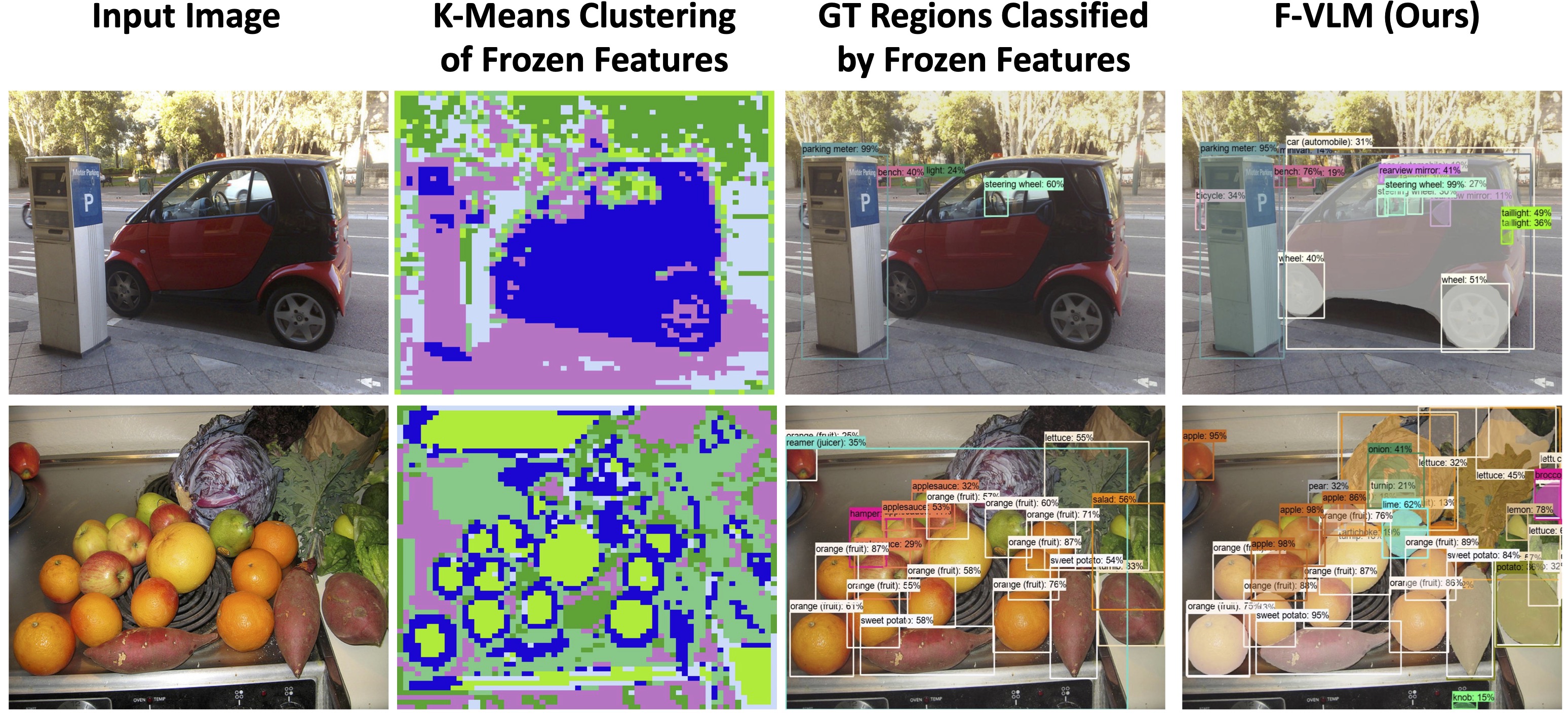}
    \caption{We explore the potential of frozen VLM (\eg, CLIP) features for open-vocabulary detection. The feature grouping reveals rich semantic and locality-sensitive information where object boundaries are nicely delineated (col. 2, see Appendix~\ref{sec:vis-clustering} for more details). The same frozen features can classify groundtruth regions well without finetuning (col. 3). Therefore, we propose to build a open-vocabulary detector on top of a frozen VLM (col. 4) without a need for knowledge distillation, detection-tailored pretraining, or weakly supervised learning. \OURS{} significantly reduces training complexity and compute requirement, and achieves the state-of-the-art performance at system level.}
    \label{fig:teaser}
    \vspace{-3mm}
\end{figure*}

Intuitively, in order to align the image content with the text description during training, VLMs may learn locality sensitive and discriminative features that are transferable to object detection. Observations in Figure~\ref{fig:teaser} support our intuition. Surprisingly, features of a frozen VLM contain rich information that are both locality sensitive for describing object shapes (col. 2) and discriminative for region classification (col. 3). This motivates us to explore using frozen VLM features for open-vocabulary detection, which entails accurate localization and classification of objects in the wild.

We propose \OURS{} -- a simple and scalable open-vocabulary detection approach built upon frozen VLMs. For localization, we simply attach a detector head to predict object regions. For open-vocabulary recognition, we apply the VLM feature pooler (\eg, a self-attention layer) on the region features from frozen backbones at test time. We train only the detector head upon a frozen VLM backbone, and combine the detection scores with the corresponding VLM predictions at test time. Our recipe reduces the training complexity of an open-vocabulary detector to below that of a standard detector, obviating the need for knowledge distillation, detection-tailored pretraining, or weakly supervised learning. By preserving the knowledge of pretrained VLMs completely, \OURS{} maintains a similar philosophy as ViTDet~\citep{li2022exploring} to decouple the detector-specific learning from the more task-agnostic vision knowledge in the backbone.

We demonstrate the efficacy of \OURS{} on LVIS~\citep{lvis}, COCO~\citep{coco} and Objects365~\citep{objects365}. Here is a summary of our contributions and observations:
\begin{itemize}
    \item We propose \OURS{} -- a simple open-vocabulary detection method upon frozen VLMs without knowledge distillation, detection-tailored pretraining, or weakly supervised learning.
    \item Despite its simplicity, \OURS{} achieves strong performance, surpassing the previous state-of-the-art on LVIS open-vocabulary detection benchmark by 6.5 mask AP$_r$ at system level and outperforming existing approaches in cross-dataset transfer (COCO, Objects365).
    \item \OURS{} shows compelling scaling behavior with consistent performance improvements by increasing the backbone capacity (\eg, +14.2 LVIS mask AP$_r$ with our largest backbone).
    \item \OURS{} has much fewer trainable parameters, allowing it to train significantly faster. Compared with a strong open-vocabulary detection method ViLD~\citep{gu2022openvocabulary}, \OURS{} not only achieves better performance, but also provides up to 200\x{} training compute savings.
\end{itemize}
We hope these findings will facilitate the research community to further explore frozen VLMs for a broader range of computer vision tasks.

\vspace{-2mm}
\section{Related Work}
\vspace{-1mm}
\paragraph{Zero-shot/Open-vocabulary visual recognition and representation learning.} Zero-shot and open-vocabulary recognition has been a long-standing problem in computer vision. Earlier works use the visual attributes to represent categories as binary codebooks and learn to predict the attributes for novel categories ~\citep{jayaraman2014zero,rohrbach2011evaluating}. DeViSE~\citep{frome2013devise} and ConSE~\citep{norouzi2013zero} pioneer to learn a joint image-text embedding space using deep learning. Many works have shown the promise of representation learning from natural language associated with images, such as image tags~\citep{chen2015webly,divvala2014learning,joulin2016learning} or text descriptions~\citep{desai2021virtex,he2017fine,sariyildiz2020learning,wang2009learning,zhong2021learning}. Recently, popular large VLMs scale up by training on billions of image-text pairs and acquire strong image-text representation by contrastive learning~\citep{radford2021clip,align,basic,zhai2021lit}. These models achieve strong zero-shot performance on many classification benchmarks and show clear benefits in scaling model capacity. 

While all the above works study image-level recognition, the focus of this paper is on the object-level understanding. Recently,~\citet{vasconcelos2022proper} has shown frozen classification models are beneficial for closed-vocabulary detection with adequate detector head capacity. In addition, a frozen VLM can serve as a teacher model and combine with self-training for zero-shot semantic segmentation~\citep{zhou2022maskclip}. In contrast, we study how to use frozen VLM directly as part of an open-vocabulary object detector.

\paragraph{Zero-Shot/Open-vocabulary object detection.} It is costly and labor-intensive to scale up data collection and annotation for large vocabulary detection. Zero-shot detection aims to alleviate the challenge by learning to detect novel categories not present in the training data. Many methods address this by aligning the image region features to category word embeddings~\citep{bansal2018zero, rahman2020improved, demirel2018zero, zheng2020background}, or synthesizing visual features with a generative model~\citep{hayat2020synthesizing,zhu2020don}. Recently, Zareian~\etal proposes the open-vocabulary detection (OVD) benchmark with a view to bridge the performance gap between ZSD and supervised learning~\citep{Zareian_2021_CVPR}. The model was first pretrained on image-caption data to recognize novel objects, and then finetuned for zero-shot detection~\citep{Zareian_2021_CVPR}. 

Following the OVD benchmark, ViLD~\citep{gu2022openvocabulary} proposes to distill the rich representation of pretrained VLM into the detector, and DetPro~\citep{du2022learning} improves upon ViLD by applying the idea of prompt optimization. RegionCLIP~\citep{zhong2021regionclip} develops a region-text pretraining strategy that leverages pretrained VLMs and image-caption data, while Detic~\citep{zhou2022detecting} jointly trains a detector with weak supervision. VL-PLM~\citep{zhao2022exploiting} explores pseudo-labeling on unlabeled data with object proposals and VLMs for OVD. GLIP~\citep{li2021grounded} formulates object detection as a phrase grounding task and pretrains on a wide variety of detection, grounding, and caption datasets for zero/few-shot object detection. Similarly, OWL-ViT~\citep{minderer2022simple} proposes to finetune pretrained vision transformers on a suite of detection/grounding datasets. All mentioned methods require training the entire detector from scratch, finetuning after detection-tailored pretraining, or training on a suite of detection/grounding datasets. In contrast, \OURS{} trains only the standard detector head upon a frozen VLM without using any of the above.

\vspace{-2mm}
\section{Method}
\vspace{-1mm}

\subsection{Overview}
\vspace{-1mm}
In this paper we address the problem of open-vocabulary object detection. At training time, the model has access to the detection labels of $C_B$ base categories, but needs to detect objects from a set of $C_N$ novel categories at test time. To make the settings more practical~\citep{Zareian_2021_CVPR}, we follow previous works and assume the availability of a pretrained vision and language model (VLM) which has learned from plenty of image-text pairs on the internet~\citep{gu2022openvocabulary}. 

Figure~\ref{fig:fvlm-system} shows the overall \OURS{} architecture. We propose to build the open-vocabulary object detector upon frozen VLMs by training only the detector head upon frozen features, which guarantees to completely preserve the open-vocabulary classification ability of pretrained VLMs. At test time, we combine the detector scores with the VLM scores to obtain open-vocabulary object detection scores. By directly using frozen pretrained models, our approach is simple and easily scalable.

\vspace{-2mm}
\subsection{Pretraining from Vision and Language Models}
\label{sec:pretrain}
Recently, Vision and Language Models (VLM) are popular because of their rich knowledge and strong representation for both visual and linguistic domains. We desire to retain their knowledge as much as possible, in order to minimize the effort/cost to adapt the VLMs for open-vocabulary detection. Following existing works~\citep{du2022learning,gu2022openvocabulary,zhong2021regionclip}, we focus on contrastively pretrained VLMs in this paper \eg~\citep{align,radford2021clip}. Contrastive VLMs typically have the image and text encoders trained jointly with a contrastive objective. We use the frozen image encoder as the detector backbone, and the text encoder for caching the text embeddings of detection dataset vocabulary offline (see Sec.~\ref{sec:text-det}).

\begin{figure*}[t]
    \centering
    \begin{subfigure}[b]{0.98\linewidth}
        \centering
    	\includegraphics[width=0.98\linewidth]{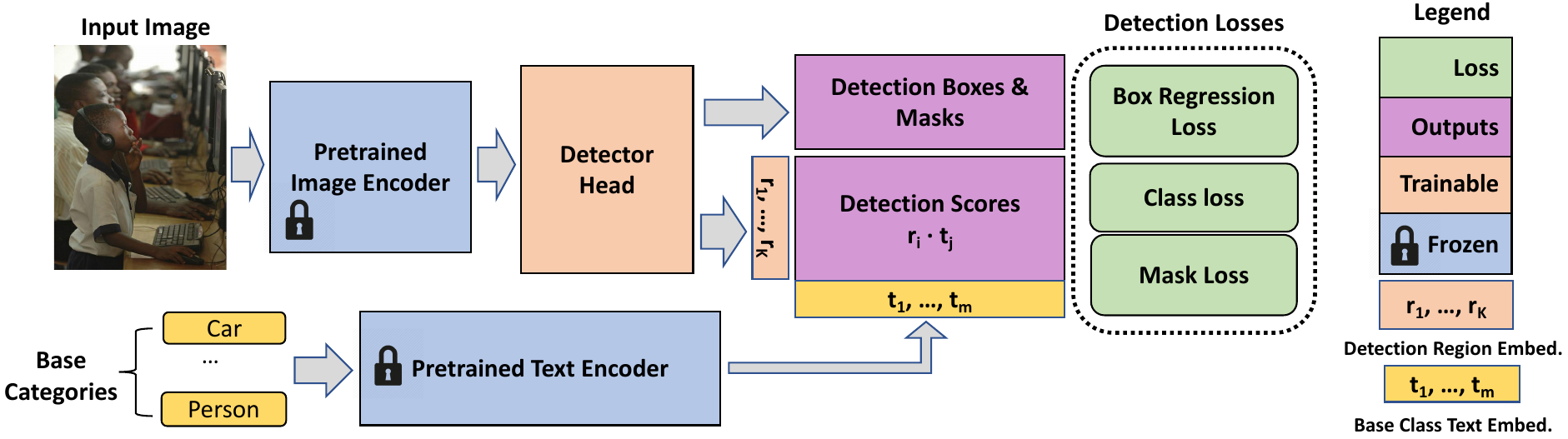}
    	\caption{\textbf{\OURS{} training architecture}. At training time, \OURS{} is simply a detector with the last classification layer replaced by base-category text embeddings. The detector head is the only trainable part of the system, which includes RPN~\citep{ren2015faster}, FPN~\citep{fpn}, and Mask R-CNN heads~\citep{he2017mask}.}
    	\label{fig:train-system}
    \end{subfigure}
    \begin{subfigure}[b]{0.98\linewidth}
        \centering
    	\includegraphics[width=0.98\linewidth]{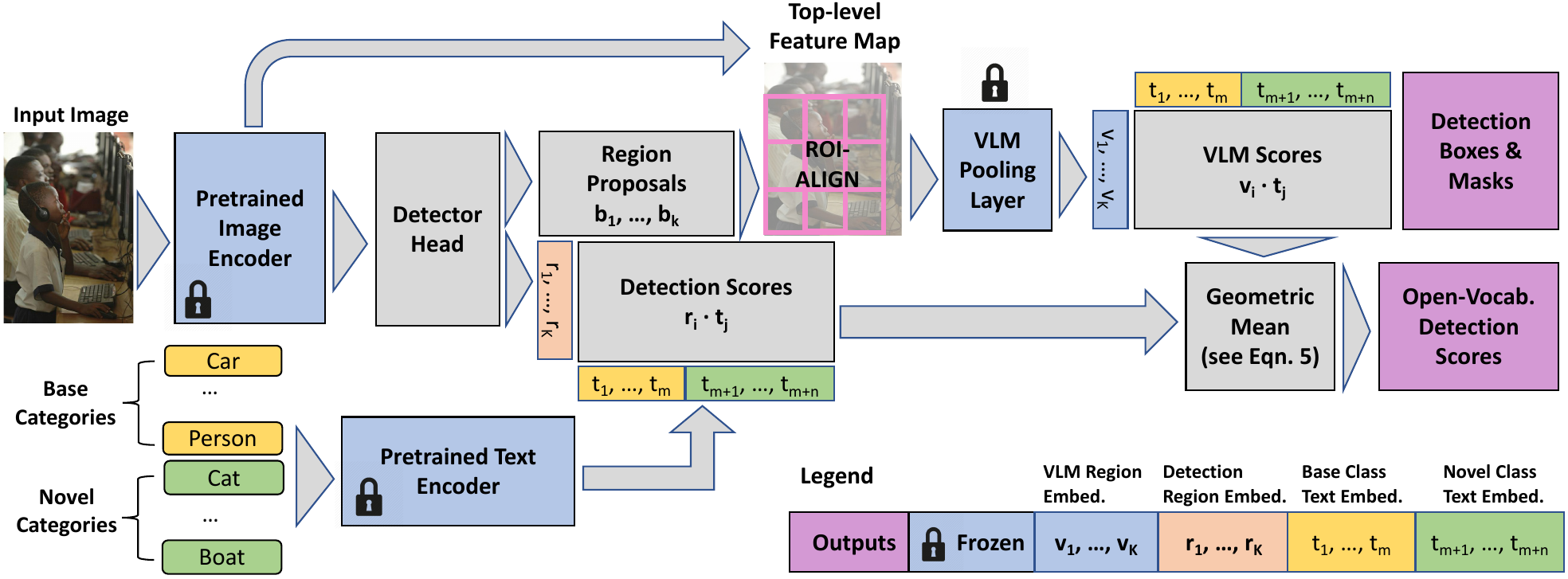}
    	\caption{\textbf{\OURS{} inference architecture}. At test time, \OURS{} uses the region proposals to crop out the top-level features of VLM backbone and compute the VLM score per region. The trained detector head provides the detection boxes and masks, while the classification scores are a combination of detection and VLM scores.}
    	\label{fig:test-system}
	\end{subfigure}
	\caption{\textbf{\OURS{} architecture.} We present both training and inference time architectures of \OURS{}, where the VLM pooling layer and detection score combination are the differences.}
	\label{fig:fvlm-system}
	\vspace{-5mm}
\end{figure*}

We consider the VLM image encoder in two parts: 1) the feature extractor $\mathcal{F}(\cdot)$, \eg ResNet-50~\citep{radford2021clip}, and 2) the last feature pooling layer $\mathcal{P}(\cdot)$, \eg an attention pooling layer~\citep{radford2021clip}. We adopt the same backbone architecture as the image feature extractor $\mathcal{F}(\cdot)$, which allows us to directly use the frozen weights and inherit the rich semantic knowledge (see Fig.~\ref{fig:train-system}). Along with the backbone initialization, we also adopt the same image pre-processing scheme as the VLM pretraining to maintain its open-vocabulary recognition ability. We use the last VLM pooling layer $\mathcal{P}(\cdot)$ for open-vocabulary recognition at test time only (see Sec.~\ref{sec:ovr}). Building upon the frozen backbone features, we adopt Mask R-CNN~\citep{he2017mask} head and feature pyramid network~\citep{fpn} as the detector head following previous works~\citep{du2022learning,gu2022openvocabulary,zhong2021regionclip}. The detector head is randomly initialized and is \textit{the only trainable component} of \OURS{}. Despite the image-level pretraining, we found empirically that the frozen VLM backbone contains adequate locality sensitive features to enable accurate downstream detection (see Appendix~\ref{sec:vis-clustering}).

\vspace{-2mm}
\subsection{Text-Embedding Region Classifier}
\label{sec:text-det}

\paragraph{Notations:} Let's define $I$ as the input image, $\mathcal{F}(I)$ the backbone features from the image encoder. Let $\mathcal{Q}(\cdot)$ be the function that yields a region embedding $\mathbf{r}_b$ from $\mathcal{F}(I)$ and a given box region proposal $b$, which involves FPN~\citep{fpn}, ROI-Align~\citep{he2017mask}, and Faster R-CNN head~\citep{ren2015faster}. We have:
\begin{equation}\label{eqn:detect-feat}
    \mathbf{r}_b = \mathcal{Q}(\mathcal{F}(I), b)
\end{equation}

Standard detectors use K-way classifier because the training and test time categories are the same. This design does not support the open-vocabulary settings which require new categories to be added at test time. To accommodate this, we replace the last fully connected layer with the text embeddings of base categories (see Fig.~\ref{fig:train-system}). At inference time, we can simply expand the text embeddings to include novel categories for open-vocabulary detection (see Fig.~\ref{fig:test-system}). An advantage of such design is that the system can generalize to the novel categories near $C_B$ in the embedding space.

To generate the text embeddings, it is critical to use the matching text encoder of the image encoder because they were pre-trained together. Apart from $C_B$, the background category is represented by a generic phrase ``background'' for compatibility with other categories. At training time, the proposals not matched to any groundtruth boxes in $C_B$ are treated as background. For each region, we compute the cosine similarity of $\mathbf{r}_b$ with the text embeddings of $C_B$ and ``background'', and apply a learnable temperature $\tau$ on the logits. The detection scores $\mathbf{z}(\mathbf{r}_b)$ are given by:
\begin{equation}\label{eqn:detect-logit}
\mathbf{z}(\mathbf{r}_b) = Softmax(\frac{1}{\tau}
    \begingroup 
    \setlength\arraycolsep{2pt}
    \begin{bmatrix}
    cos(\mathbf{r}_b, \mathbf{t}_{bg}), & 
    cos(\mathbf{r}_b, \mathbf{t}_1), &
    \cdots, & 
    cos(\mathbf{r}_b, \mathbf{t}_{|C_B|})
    \end{bmatrix})
    \endgroup    
\end{equation}
where $cos(\mathbf{a}, \mathbf{b}) = \mathbf{a}^\top \mathbf{b} / (\|\mathbf{a}\|\|\mathbf{b}\|)$, and $\mathbf{t}_i$ denotes the text embeddings of class $i$. We apply the standard softmax cross entropy loss on the logits (see Fig.~\ref{fig:train-system}). At test time, we keep the ``background'' category and expand the text embeddings from $C_B$ to $C_B \cup C_N$ for open-vocabulary detection. Similar designs have been used by previous works~\citep{Zareian_2021_CVPR,gu2022openvocabulary}.

\vspace{-1mm}
\subsection{Open-Vocabulary Recognition}
\label{sec:ovr}

The ability to perform open-vocabulary recognition at region level is integral to \OURS{}. Since the backbone features are frozen, they do not overfit to the base categories and can be directly cropped for region-level classification. \OURS{} performs this open-vocabulary classification only at test time.

To obtain the features for a region $b$, we apply the VLM pooling layer $\mathcal{P}(\cdot)$ on the cropped backbone output features  $\mathcal{F}(I)$ (see Sec.~\ref{sec:pretrain} for notations). Because the pooling layer requires fixed-size inputs, \eg 7x7 for R50~\citep{radford2021clip}, we crop and resize the region features with ROI-Align $\mathcal{R}(\cdot)$~\citep{he2017mask} (see Fig.~\ref{fig:test-system}). Unlike existing works~\citep{gu2022openvocabulary, du2022learning}, we do not crop and resize the RGB image regions and cache their embeddings in a separate offline process, but train the detector head in one stage. This is simpler and more space-efficient. In addition, we do not crop VLM region features with $\mathcal{R}(\cdot)$ during training because the backbone features are frozen. Using the notations from~\eqref{eqn:detect-feat}, we obtain the VLM region embedding $\mathbf{v}_b$ by:
\begin{equation}
    \mathbf{v}_b = \mathcal{P}(\mathcal{R}(\mathcal{F}(I), b))
\end{equation}

where $b$ denotes the box region and $\mathbf{v}_b$ corresponds to $v_1,...,v_k$ in Fig.~\ref{fig:test-system}. Note $\mathcal{R}(\cdot)$ is used at test time only. Similar to~\eqref{eqn:detect-logit}, we compute the VLM scores by cosine similarity as follows:
\begin{equation}\label{eqn:vlm-logit}
\mathbf{w}(\mathbf{v}_b) = Softmax(\frac{1}{T}
    \begingroup
    \setlength\arraycolsep{2pt}
    \begin{bmatrix}
    cos(\mathbf{v}_b, \mathbf{t}_{bg}), & 
    cos(\mathbf{v}_b, \mathbf{t}_1), &
    \cdots, & 
    cos(\mathbf{v}_b, \mathbf{t}_{|C_{B \cup N}|})
    \end{bmatrix})
    \endgroup    
\end{equation}
where $T$ is a fixed temperature and the text embeddings include both the $C_B$ and $C_N$ at inference time (see Fig. \ref{fig:test-system}). We use a fixed temperature to adjust the scale of VLM scores relative to the detection scores in~\eqref{eqn:detect-logit}. In the special case when the region $b$ is equal to the whole image, the VLM scores $\mathbf{w}(\mathbf{v}_b)$ becomes equivalent to the zero-shot image classification scores. 

Despite never being trained on regions, the cropped region features of $\mathcal{F}(\cdot)$ maintain good open-vocabulary recognition ability. However, we observe the cropped region features are not sensitive enough to the localization quality of the regions, \ie a loosely vs tightly localized box both have similar features. This may be good for classification, but is problematic for detection because we need the detection scores to reflect localization quality as well. To remedy this, we apply the geometric mean to combine the VLM scores $w(\mathbf{v}_b)_i$  in~\eqref{eqn:vlm-logit} with the detection scores $z(\mathbf{r}_b)_i$ in~\eqref{eqn:detect-logit} for each region $b$ and category $i$. The final detection scores $s(\mathbf{r}_b)_i$ are given by:
\begin{equation}\label{eqn:combine-score}
s(\mathbf{r}_b)_i = \begin{cases}
    z(\mathbf{r}_b)_i^{(1-\alpha)} \cdot w(\mathbf{v}_b)_i^\alpha & \text{if } i \in C_B\\
    z(\mathbf{r}_b)_i^{(1-\beta)} \cdot w(\mathbf{v}_b)_i^\beta & \text{if } i \in C_N
\end{cases}
\end{equation}
where $\alpha$, $\beta \in [0, 1]$ control the VLM score weights for base/novel categories, and the background score comes directly from the detector \ie, $s(\mathbf{r}_b)_0 = z(\mathbf{r}_b)_0$. Compared to the ensemble system in~\citep{gu2022openvocabulary}, our design is simpler without a need for knowledge distillation or double Faster R-CNN heads. We show ablations of different score fusion designs in Appendix~\ref{sec:score-fuse}.

\subsection{Open-Vocabulary Localization}
\label{sec:ovloc}

How to localize and separate the novel objects from the background is an important problem in open-vocabulary detection. Standard detectors are not designed for localizing novel objects because most of them apply class-specific localization, including the box regression and mask prediction heads, \eg, Mask R-CNN~\citep{he2017mask}. Inspired by the learned objectness~\citep{oln, Kuo_2015_ICCV, wang2020leads}, we use \textit{class-agnostic} box regression and mask prediction heads instead. That is, for each region proposal, we predict one box and one mask for all categories, rather than one per category. This simple change allows us to localize novel objects in the open-vocabulary settings. We note that \OURS{} framework is not specific to the choice of Mask R-CNN detector head and other models can potentially be applied as well \eg~\citep{carion2020end,redmon2016you}. We choose Mask R-CNN per existing works~\citep{gu2022openvocabulary,Zareian_2021_CVPR,zhong2021regionclip}.

\vspace{-3mm}
\section{Experiments}
\vspace{-2mm}

\paragraph{Implementation Details.}
\label{sec:impl}
We choose Mask R-CNN~\citep{he2017mask} with feature pyramid network~\citep{fpn} as our detector head throughout the paper. The head design follows~\citep{ghiasi2021simple,gu2022openvocabulary}. We train the model for 46.1k iterations with 1024x1024 image size, large scale jittering~\citep{ghiasi2021simple}, batch size 256, weight decay 1e-4, momentum 0.9, and an initial learning rate 0.36. For the score combination, we use $\alpha=0.35$ and $\beta=0.65$ in~\eqref{eqn:combine-score}. We use a maximum of $300$ detections per image, and set temperature $T=0.01$ in~\eqref{eqn:vlm-logit}. We use CLIP~\citep{radford2021clip} prompt templates and take the average text embeddings of each category. Please refer to Appendix~\ref{sec:appendix-impl} for a full list of hyper-parameter configurations.
\vspace{-2mm}
\subsection{Open-Vocabulary Detection Benchmark}

\paragraph{LVIS Benchmark.} We evaluate our approach on the LVIS dataset~\citep{lvis} which contains a large and diverse set of 1203 object categories suitable for open-vocabulary detection. Following the existing works~\citep{gu2022openvocabulary,zhong2021regionclip}, we treat the frequent and common categories as the base categories $C_B$ for training, and hold out the rare categories as novel categories $C_N$ for testing. Mask AP$_r$ is the main metric we benchmark on. To ensure reproducibility, we report the mean of 5 independent runs following the protocol of~\citep{gu2022openvocabulary} and the best practice of LVIS challenge~\citep{lvis}. For fair comparison, we adopt the same Mask R-CNN head architecture as~\citep{gu2022openvocabulary} and use the same large scale jittering recipe~\citep{ghiasi2021simple}.

Table~\ref{table:main_results} presents our results on LVIS. In the R50 comparison, \OURS{} ranks second among the other alternatives based on knowledge distillation, pretraining, or joint training with weak supervision. The leading DetPro~\citep{du2022learning} shows the effectiveness of prompt optimization~\citep{zhou2022cocoop} which can potentially benefit \OURS{} too. In the system-level comparison, we observe the performance of \OURS{} scales up nicely with frozen model capacity, even though the amount of trainable parameters is fixed. Our best model achieves 32.8 AP$_r$, which is +14.2 AP$_r$ from the R50 baseline and the best published results on this benchmark to our knowledge. Compared to the best existing approach (ViLD-EN-B7), we outperform by 6.5 mask AP$_r$ on the novel categories (and +5.6 overall mask AP). We provide additional results using standard 1x/3x training recipes~\citep{wu2019detectron2} in Appendix~\ref{sec:standard-recipe}, where \OURS{} shows similarly strong performance on shorter recipes and smaller batch size by using frozen backbones.

\begin{table}[h]
\caption{
\textbf{LVIS Open-Vocabulary Object Detection Benchmark}. \OURS{} outperforms the best existing approach by 6.5 mask AP on novel categories. All methods use the same instance-level supervision from LVIS~\citep{lvis} base categories, CLIP~\citep{radford2021clip} pretraining, and fixed prompt templates unless noted otherwise. $^{\dagger}$: Pretraining with CC-3M~\citep{ccdataset}. $^{\ddagger}$: Prompt optimization~\citep{zhou2022cocoop} and SoCo pretraining~\citep{SoCo}. $^{\ast}$: Joint training with IN-21k~\citep{deng2009imagenet}. $^{\star}$: ALIGN model~\citep{align}.}

\label{table:main_results}
\centering
{\footnotesize
\begin{tabular}{lllccl>{\color{gray}}l}
\toprule
Backbone \textcolor{gray}{(\# Params)} & \makecell{Pretrained \\CLIP} & Method  & Distill &  \makecell{Trainable\\Backbone} & AP$_r$ & AP\\
\midrule
\multicolumn{4}{l}{R50 Comparison:} & & &\\
\midrule
R50 & ViT-B/32  & ViLD~\citep{gu2022openvocabulary} & \cmark & \cmark & 16.1 & 22.5\\
R50 & ViT-B/32  & ViLD-Ens.~\citep{gu2022openvocabulary} & \cmark & \cmark & 16.6 & 25.5\\
R50 & ViT-B/32  & DetPro~\citep{du2022learning}$^{\ddagger}$ & \cmark & \cmark & 19.8 & 25.9 \\ 
R50 & ViT-B/32 & Detic-ViLD~\citep{zhou2022detecting}$^{\ast}$ & \xmark & \cmark & 17.8 & 26.8 \\
R50 & R50  & RegionCLIP~\citep{zhong2021regionclip}$^{\dagger}$ & \cmark & \cmark & 17.1 & 28.2 \\
R50 & R50 & \OURS{} (Ours) & \xmark & \xmark & 18.6 & 24.2 \\
\midrule
\multicolumn{4}{l}{System-level Comparison:} & & & \\
\midrule
R152 \textcolor{gray}{(60M)} & ViT-B/32  & ViLD~\citep{gu2022openvocabulary} & \cmark & \cmark & 18.7 & 23.6 \\
R152 \textcolor{gray}{(60M)} & ViT-B/32  & ViLD-Ens.~\citep{gu2022openvocabulary} & \cmark & \cmark & 18.7 & 26.0 \\
EN-B7 \textcolor{gray}{(67M)} & ViT-L/14 & ViLD-Ens.~\citep{gu2022openvocabulary} & \cmark & \cmark & 21.7 & 29.6 \\
EN-B7 \textcolor{gray}{(67M)} & EN-B7$^{\star}$ & ViLD-Ens.~\citep{gu2022openvocabulary} & \cmark & \cmark & 26.3 & 29.3 \\
R50 \textcolor{gray}{(26M)} & ViT-B/32 &  DetPro-Cascade~\citep{du2022learning}$^{\ddagger}$ & \cmark & \cmark & 20.0 & 27.0  \\
R50 \textcolor{gray}{(26M)} & ViT-B/32 & Detic-CN2~\citep{zhou2022detecting}$^{\ast}$ & \xmark & \cmark & 24.6 & 32.4 \\
R50x4 \textcolor{gray}{(87M)} & R50x4 & RegionCLIP~\citep{zhong2021regionclip}$^{\dagger}$ & \cmark & \cmark & 22.0 & 32.3 \\
ViT-L/14 \textcolor{gray}{(303M)} & ViT-L/14 & OWL-ViT~\citep{minderer2022simple} & \xmark & \cmark & 25.6 & 34.7 \\
R50x4 \textcolor{gray}{(87M)} & R50x4  & \OURS{} (Ours) &  \xmark & \xmark & 26.3 & 28.5 \\
R50x16 \textcolor{gray}{(167M)} & R50x16 & \OURS{} (Ours) &  \xmark & \xmark & 30.4 & 32.1 \\
R50x64 \textcolor{gray}{(420M)} & R50x64 & \OURS{} (Ours) &  \xmark & \xmark & \textbf{32.8} & 34.9 \\
\bottomrule
\end{tabular}
\vspace{-2mm}
}
\end{table}

\paragraph{COCO Benchmark.} Many existing works on zero-shot detection~\citep{bansal2018zero} and open-vocabulary detection~\citep{Zareian_2021_CVPR,gu2022openvocabulary,zhong2021regionclip} benchmark on COCO. This setup divides COCO vocabulary into 48 base categories for training and 17 novel categories for testing. We follow the standard practice and report results in the generalized detection settings without instance segmentation. The main metric is AP50 of novel categories. Similar to LVIS, we report the mean of 5 independent runs to ensure reproducibility.

Due to the smaller number of training categories, we observe a tendency to overfit when we re-use the same LVIS training recipe. F-VLM does not rely on additional objectives \eg knowedlge distillation or weak supervision to counter-balance overfitting. We therefore reduce the training epoch, background weight, and increase the weight decay to mitigate this. Please refer to Appendix~\ref{sec:appendix-impl} for a full list of hyper-parameters.

Table~\ref{table:coco} shows that \OURS{} is very competitive among the published results. Compared to the leading RegionCLIP~\citep{zhong2021regionclip} which uses additional caption pretraining, \OURS{} directly uses a frozen CLIP backbone. In fact, \OURS{} significantly surpasses the CLIP-R50 pretrained version of RegionCLIP, which does not leverage pretraining on caption data. Compared to other approaches, \OURS{} offers better performance without the use of detection-tailored pretraining, weakly supervised learning, or knowledge distillation.

\begin{table}[h]
\caption{
\textbf{COCO Open-Vocabulary Object Detection Benchmark.} \OURS{} is very competitive with the other methods trained with various sources. All methods use the ResNet50 backbone~\citep{he2016deep,radford2021clip}. RegionCLIP additionally use COCO Captions$^{\dagger}$~\citep{coco} or CC3M$^\ddagger$~\citep{ccdataset} for pretraining. $^{\star}$: CLIP initialization without region-level pretraining. $^{\ast}$: Joint training with COCO captions.
}
\label{table:coco}
\centering
{\footnotesize
\begin{tabular}{lcc>{\color{gray}}c}
\toprule
Method  & Training source & Novel AP &  AP\\
\midrule
WSDDN~\citep{WSDDN}       & \multirow{2}{*}{image-level labels in $C_B \cup C_N$} & 19.7 & 19.6 \\
Cap2Det~\citep{Cap2Det}     &       & 20.3 & 20.1 \\
\midrule
ZSD~\citep{bansal2018zero}          &   \multirow{3}{*}{instance-level labels in $C_B$} & 0.31 & 24.9 \\
DELO~\citep{zhu2020don}        &       & 3.41 & 13.0 \\
PL~\citep{rahman2020improved}  &       & 4.12 & 27.9 \\
\midrule
OVR-CNN~\citep{Zareian_2021_CVPR} & \makecell{image captions in $C_B \cup C_N$\\instance-level labels in $C_B$}  & 22.8 & 39.9\\
\midrule
CLIP-RPN~\citep{gu2022openvocabulary} &  \multirow{7}{*}{\makecell{CLIP image-text pairs\\instance-level labels in $C_B$}}  & 26.3 & 27.8\\
ViLD~\citep{gu2022openvocabulary} & & 27.6 & 51.3 \\
Detic$^{\ast}$~\citep{zhou2022detecting} & & 27.8 & 45.0 \\
RegionCLIP$^\ddagger$ ~\citep{zhong2021regionclip} &   &  \textbf{31.4} & 50.4 \\
RegionCLIP$^{\dagger}$ ~\citep{zhong2021regionclip} &   &  26.8 & 47.5 \\
RegionCLIP$^{\star}$ ~\citep{zhong2021regionclip} & & 14.2 & 42.7 \\
\OURS{} (Ours) & & 28.0 & 39.6 \\
\bottomrule
\end{tabular}
\vspace{-4mm}
}
\end{table}

\paragraph{Training Resource Benchmark.} We explore the benefits of frozen VLMs in terms of training resource savings. We benchmark with ViLD~\citep{gu2022openvocabulary} as it is most comparable to \OURS{}. Both adopt the same Mask R-CNN head configuration and training recipe~\citep{ghiasi2021simple}, and neither require detection-tailored pretraining. We follow ViLD and compare the training cost on TPUv3 cores on the same batch size. The data about ViLD training time and resource use is obtained directly from the authors~\citep{gu2022openvocabulary}. To keep the benchmark simple, we assume the pretrained VLMs are given and exclude their training costs from the comparison. For \OURS{}, we use the R50x64 backbone and report the average over 5 independent runs.

\begin{table*}[h]
	\caption{\textbf{Training Resource Benchmark.} We report LVIS mask AP$_r$ to show the performance vs training cost trade-off. \OURS{} can outperform ViLD~\citep{gu2022openvocabulary} with 226\x{} less compute.}
	\vspace{-1.5mm}
\small
\centering
    \begin{tabular}{l|c|cc|cc}
    \toprule
    Method & Mask AP$_r$ & \#Iters & Epochs & \makecell{Training Cost \\ (Per-Core-Hour)} & \makecell{Training Cost\\Savings} \\
    \midrule
    ViLD-EN-B7~\citep{gu2022openvocabulary} & 26.3 & 180k & 460 & 8000 & 1$\times$ \\
    \midrule
    \OURS{} (Ours) & 32.8 & 46.1k & 118  & 565 & 14$\times$ \\
    \OURS{} (Ours) & 31.0 & 5.76k & 14.7 & 71 & 113$\times$ \\
    \OURS{} (Ours) & 27.7 & \textbf{2.88k} & \textbf{7.4} & \textbf{35} & \textbf{226$\times$} \\
    \bottomrule
    \end{tabular}
    \vspace{-2mm}
	\label{table:compute}
\end{table*}

Table \ref{table:compute} shows that \OURS{} can achieve top performance with much less compute. Compared to the  state-of-the-art (ViLD-EN-B7) at system level, \OURS{} can achieve better performance with only 7.4 epochs of training, which is 226\x{} more compute-efficient and 57$\times$ faster in wall clock time. We believe the efficiency gain arises from the frozen backbone, which substantially simplifies the learning process. This is orthogonal to the detection-tailored pretraining used by existing works to speed up training~\citep{Zareian_2021_CVPR,zhong2021regionclip}. Apart from resource savings, \OURS{} has potential for substantial memory savings at training time by running the backbone in inference mode (see Appendix~\ref{sec:memory-use} for more details). \OURS{} system runs almost as fast as a standard detector~\citep{he2017mask} at inference time, because the only addition is a single attention pooling layer~\citep{radford2021clip} on the detected region features (see Fig.~\ref{fig:test-system}). 

\paragraph{Transfer Detection Benchmark.} We explore the potential of \OURS{} as a general-purpose detector for different data sources with a view to move towards non dataset-specific detection. \OURS{} trained on one dataset can be directly applied to another by swapping out the vocabulary without any finetuning, \eg, replacing the 1203 LVIS categories with COCO 80 categories. The models we use are trained on LVIS base categories and tested on COCO and Objects365-v1 validation splits following the transfer setup of ViLD~\citep{gu2022openvocabulary}. Since COCO and Objects365 have smaller vocabularies than LVIS, category and image overlaps are hard to avoid. We calculate the vocabulary overlap between COCO/Objects365 and LVIS base categories to be $91\%$ and $63\%$ respectively. Please refer to Appendix~\ref{sec:vocab-overlap} for more discussion about this benchmark.

Table~\ref{table:transfer} presents the results in comparison with prior works and supervised baselines. We observe the performance of \OURS{} improves steadily as we scale up the frozen model capacity. On Objects365/COCO, the best \OURS{} outperforms existing works ViLD by +3.2/+5.9 and DetPro by +4.9/+5.6, closing the gap with a supervised model on COCO (-33\%) and Objects365 (-40\%). The results are reported in Box AP averaged over 5 runs. There is no longer distinction between base and novel categories in the transfer setting, so we assume all categories are novel and use $\beta$ alone to combine detection and VLM scores in~\eqref{eqn:combine-score} (see Sec.~\ref{sec:ovr}). We observe that only the detection scores are needed ($\beta=0$) for COCO, while the optimal $\beta=0.3$ to $0.4$ on Objects365.

\vspace{-2mm}
\begin{table*}[h]
	\caption{\textbf{Transfer detection of \OURS{}}. We evaluate LVIS-trained \OURS{} on COCO and Objects365 without finetuning. \OURS{} demonstrates strong scaling property with a gain of +7.3/+5.8 AP on COCO/Objects365 by increasing backbone capacity.}
	\vspace{-1mm}
\small
\centering
    \begin{tabular}{l|ccc|ccc}
    \toprule
      \multirowcell{2}[0ex][l]{Method} & \multicolumn{3}{c|}{COCO} & \multicolumn{3}{c}{Objects365}\\
    & AP & AP$_{50}$ & AP$_{75}$ & AP & AP$_{50}$ & AP$_{75}$ \\
    \midrule
    Supervised~\citep{gu2022openvocabulary} & 46.5 & 67.6 & 50.9 &  25.6 & 38.6 & 28.0 \\
    \midrule
    ViLD-R50~\citep{gu2022openvocabulary}      & 36.6 & 55.6 & 39.8 & 11.8 & 18.2 & 12.6 \\
    DetPro-R50~\citep{du2022learning} & 34.9 & 53.8 & 37.4 & 12.1 & 18.8 & 12.9 \\
    \OURS-R50 (Ours)      &  32.5 & 53.1 & 34.6 & 11.9 & 19.2 & 12.6 \\
    \OURS-R50x4 (Ours)    &  36.0 & 57.5 & 38.7 & 14.2 & 22.6 & 15.2 \\
    \OURS-R50x16 (Ours)    &  37.9 & 59.6 & 41.2 & 16.2 & 25.3 & 17.5 \\
    \OURS-R50x64 (Ours)    &  \textbf{39.8} & \textbf{61.6} & \textbf{43.8} & \textbf{17.7} & \textbf{27.4} & \textbf{19.1} \\
    \bottomrule
    \end{tabular}
	\label{table:transfer}
\end{table*}

\vspace{-4mm}
\subsection{Analyses and Visualization}
\vspace{-1mm}

\label{sec:analysis-and-ablations}

\paragraph{Ablations.} 

We present ablation studies on backbone finetuning, score fusion design/parameters, feature pyramid capacity, and background weight in Appendix~\ref{sec:ablations}. Here we summarize our findings.

In the exploration of finetuning vs frozen backbone (see Table~\ref{tab:backbone-finetuning}), we discover that finetuning improves the standard detection (base categories) but slightly hurts the open-vocabulary detection (novel categories). It remains an open question whether more sophisticated finetuning strategies can benefit open-vocabulary detection.

In the score fusion studies (see Table~\ref{tab:score-fuse}), we observe that geometric mean is  significantly better than the arithmetic mean (+8 AP$_r$). It is likely because the geometric mean requires a high-scoring region to have good detection and VLM scores simultaneously, whereas the arithmetic mean may favor regions with high detection or VLM scores. In Table~\ref{tab:score-novel} and Table~\ref{tab:score-base}, we study the score fusion weights and observe that $\beta=0.65$ and $\alpha=0.35$ are most beneficial (see~\eqref{eqn:combine-score}). Neither detection nor VLM scores alone are sufficient as $\beta=0,1$ both yield sub-optimal performances. In Table~\ref{tab:temperature}, we study the temperature in~\eqref{eqn:vlm-logit} and find the optimal $T = 10^{-2}$ is much smaller than the value of learnable $\tau \approx 1.0$ at the end of training (see~\eqref{eqn:detect-logit}). This highlights the need to use a separate $T$ for VLM scores instead of using $\tau$ for both detection and VLM scores.

In Table~\ref{tab:large-fpn}, we explore the effects of increasing feature pyramid capacity to enhance the representation learned upon the frozen backbone features. Our results show that larger pyramid benefits standard detection (base categories) without compromising the open-vocabulary detection (novel categories). In Table~\ref{tab:bg-weight}, we study the influence of background weights by~\citet{Zareian_2021_CVPR,zhong2021regionclip} on open-vocabulary detection and found it to help slightly (0.1 to 0.5 AP$_r$). Please refer to Appendix~\ref{sec:ablations} for the full experimental results.

\begin{figure*}[h]
    \vspace{-1mm}
	\includegraphics[width=0.98\linewidth]{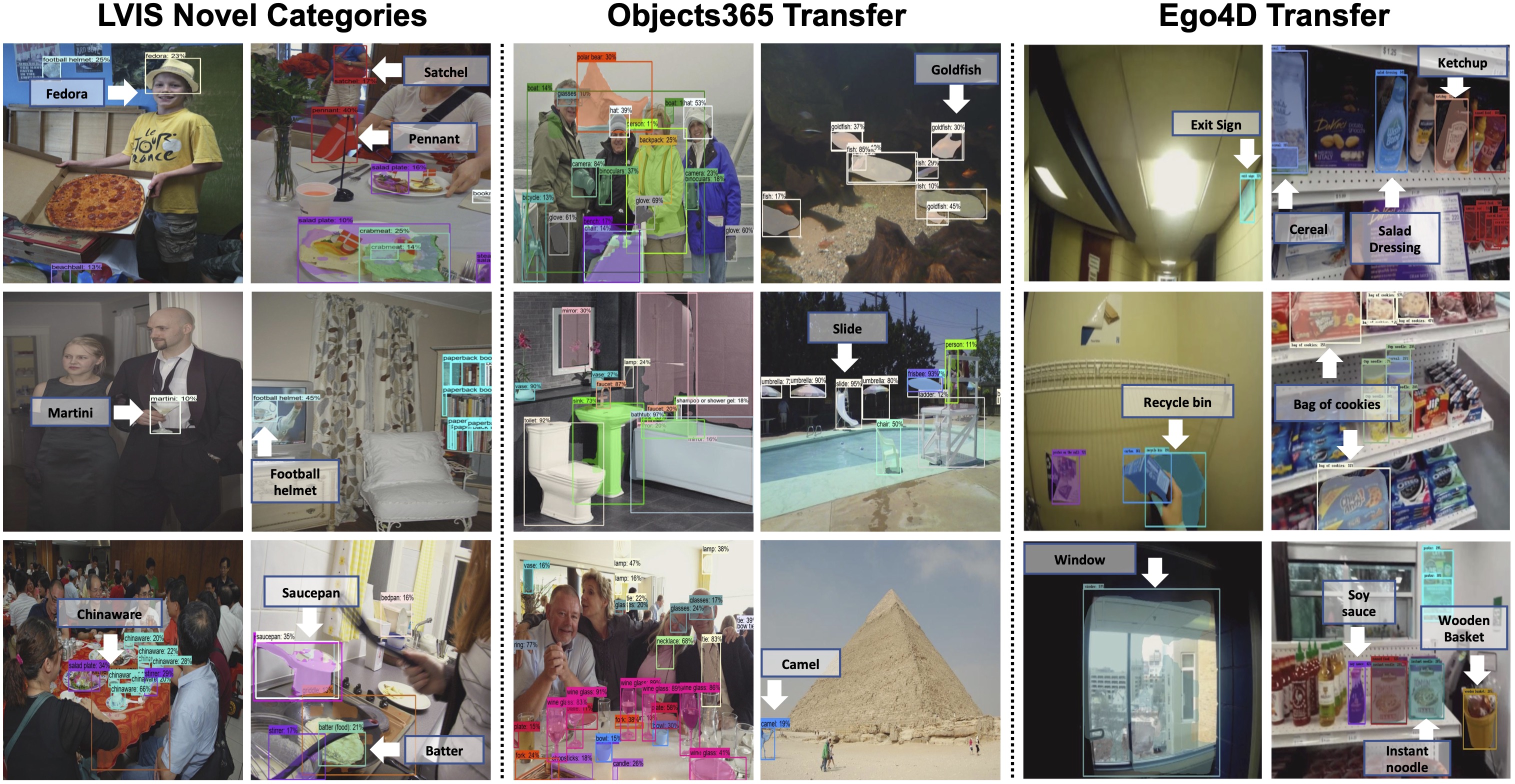}
	\vspace{-1mm}
	\caption{\textbf{\OURS{} open-vocabulary and transfer detections}. 1-2nd col.: Open-vocabulary detection on LVIS. We only show the novel categories for clarity. 2-4th col.: Transfer detection on Objects365. 4-6th col.: Transfer detection on Ego4D. Novel categories detected: \textit{fedora, martini, pennant, football helmet} (LVIS); \textit{camel, slide, goldfish} (Objects365); \textit{exit sign, recycle bin, window, soy sauce, wooden basket, cereal, bag of cookies, instant noodle, salad dressing, ketchup} (Ego4D).}
	\label{fig:vis-det}
	\vspace{-4mm}
\end{figure*}

\paragraph{Detection Visualization.} Figure~\ref{fig:vis-det} visualizes \OURS{} on open-vocabulary detection and transfer detection tasks. The transfer detection is done by replacing the dataset vocabulary without finetuning. On LVIS and Objects365~\citep{objects365}, \OURS{} correctly detects both novel and common objects. Please see Appendix~\ref{sec:more-vis} for more details and visualization. 

A key benefit of open-vocabulary detection is to test on out-of-distribution data with categories given by users on the fly. Thus, we apply \OURS{} to Ego4D~\citep{Grauman_2022_CVPR}, a real-world ego-centric application. Despite the large domain shift, \OURS{} is able to detect many novel and common objects. Please see Appendix~\ref{sec:ego4d} for more details and visualization.

\vspace{-2.5mm}
\section{Conclusion}
\vspace{-2.5mm}

We present \OURS{} -- a simple open-vocabulary detection method built upon \textit{frozen} VLMs without a need for knowledge distillation, detection-tailored pretraining, or weakly supervised learning. \OURS{} offers significant training speedup and compute savings, achieves the new state-of-the-art on LVIS benchmark at system level, and shows very competitive transfer detection. We hope this study can help the community explore frozen VLMs for a wider range of vision tasks.

\section{Reproducibility statement}
We plan to open source the code for reproducibility. We have provided the model, experimental and implementation details in the paper or the supplemental materials (Section~\ref{sec:impl} and Section~\ref{sec:appendix-impl}).
The CLIP model~\citep{radford2021clip}, Mask R-CNN model~\citep{he2017mask}, and all datasets used~\citep{coco,lvis,objects365,Grauman_2022_CVPR} in this work are publicly available.

\section{Ethics statement}
We demonstrate new capabilities in detecting previously unseen categories of objects, and particularly on challenging benchmarks and transfer settings. Our models utilize the rich information embedded in Vision-Language Models, which may reinforce deficiencies and biases in the internet data and propagate potentially harmful biases or stereotypes. The models we trained are used here for evaluation/benchmark purposes and need more rigorous probing for bias, fairness, \etc, before using them for any other purpose. 

\section{Acknowledgments}
We thank our colleagues at Google Research for their advice and helpful discussion.

\bibliography{iclr2023_conference}

\begin{thebibliography}{59}
\providecommand{\natexlab}[1]{#1}
\providecommand{\url}[1]{\texttt{#1}}
\expandafter\ifx\csname urlstyle\endcsname\relax
  \providecommand{\doi}[1]{doi: #1}\else
  \providecommand{\doi}{doi: \begingroup \urlstyle{rm}\Url}\fi

\bibitem[Bansal et~al.(2018)Bansal, Sikka, Sharma, Chellappa, and
  Divakaran]{bansal2018zero}
Ankan Bansal, Karan Sikka, Gaurav Sharma, Rama Chellappa, and Ajay Divakaran.
\newblock Zero-shot object detection.
\newblock In \emph{ECCV}, 2018.

\bibitem[Bilen \& Vedaldi(2016)Bilen and Vedaldi]{WSDDN}
Hakan Bilen and Andrea Vedaldi.
\newblock Weakly supervised deep detection networks.
\newblock In \emph{CVPR}, 2016.

\bibitem[Carion et~al.(2020)Carion, Massa, Synnaeve, Usunier, Kirillov, and
  Zagoruyko]{carion2020end}
Nicolas Carion, Francisco Massa, Gabriel Synnaeve, Nicolas Usunier, Alexander
  Kirillov, and Sergey Zagoruyko.
\newblock End-to-end object detection with transformers.
\newblock In \emph{ECCV}, 2020.

\bibitem[Chen \& Gupta(2015)Chen and Gupta]{chen2015webly}
Xinlei Chen and Abhinav Gupta.
\newblock Webly supervised learning of convolutional networks.
\newblock In \emph{ICCV}, 2015.

\bibitem[Demirel et~al.(2018)Demirel, Cinbis, and
  Ikizler-Cinbis]{demirel2018zero}
Berkan Demirel, Ramazan~Gokberk Cinbis, and Nazli Ikizler-Cinbis.
\newblock Zero-shot object detection by hybrid region embedding.
\newblock In \emph{BMVC}, 2018.

\bibitem[Deng et~al.(2009)Deng, Dong, Socher, Li, Li, and
  Fei-Fei]{deng2009imagenet}
Jia Deng, Wei Dong, Richard Socher, Li-Jia Li, Kai Li, and Li~Fei-Fei.
\newblock Imagenet: A large-scale hierarchical image database.
\newblock In \emph{CVPR}, 2009.

\bibitem[Desai \& Johnson(2021)Desai and Johnson]{desai2021virtex}
Karan Desai and Justin Johnson.
\newblock Virtex: Learning visual representations from textual annotations.
\newblock In \emph{CVPR}, 2021.

\bibitem[Divvala et~al.(2014)Divvala, Farhadi, and
  Guestrin]{divvala2014learning}
Santosh~K Divvala, Ali Farhadi, and Carlos Guestrin.
\newblock Learning everything about anything: Webly-supervised visual concept
  learning.
\newblock In \emph{CVPR}, 2014.

\bibitem[Du et~al.(2022)Du, Wei, Zhang, Shi, Gao, and Li]{du2022learning}
Yu~Du, Fangyun Wei, Zihe Zhang, Miaojing Shi, Yue Gao, and Guoqi Li.
\newblock Learning to prompt for open-vocabulary object detection with
  vision-language model.
\newblock In \emph{CVPR}, 2022.

\bibitem[Frome et~al.(2013)Frome, Corrado, Shlens, Bengio, Dean, Ranzato, and
  Mikolov]{frome2013devise}
Andrea Frome, Greg~S Corrado, Jon Shlens, Samy Bengio, Jeff Dean, Marc'Aurelio
  Ranzato, and Tomas Mikolov.
\newblock Devise: A deep visual-semantic embedding model.
\newblock In \emph{NeurIPS}, 2013.

\bibitem[Ghiasi et~al.(2021)Ghiasi, Cui, Srinivas, Qian, Lin, Cubuk, Le, and
  Zoph]{ghiasi2021simple}
Golnaz Ghiasi, Yin Cui, Aravind Srinivas, Rui Qian, Tsung-Yi Lin, Ekin~D Cubuk,
  Quoc~V Le, and Barret Zoph.
\newblock Simple copy-paste is a strong data augmentation method for instance
  segmentation.
\newblock In \emph{CVPR}, 2021.

\bibitem[Grauman et~al.(2022)Grauman, Westbury, Byrne, Chavis, Furnari,
  Girdhar, Hamburger, Jiang, Liu, Liu, Martin, Nagarajan, Radosavovic,
  Ramakrishnan, Ryan, Sharma, Wray, Xu, Xu, Zhao, Bansal, Batra, Cartillier,
  Crane, Do, Doulaty, Erapalli, Feichtenhofer, Fragomeni, Fu, Gebreselasie,
  Gonz\'alez, Hillis, Huang, Huang, Jia, Khoo, Kol\'a\v{r}, Kottur, Kumar,
  Landini, Li, Li, Li, Mangalam, Modhugu, Munro, Murrell, Nishiyasu, Price,
  Ruiz, Ramazanova, Sari, Somasundaram, Southerland, Sugano, Tao, Vo, Wang, Wu,
  Yagi, Zhao, Zhu, Arbel\'aez, Crandall, Damen, Farinella, Fuegen, Ghanem,
  Ithapu, Jawahar, Joo, Kitani, Li, Newcombe, Oliva, Park, Rehg, Sato, Shi,
  Shou, Torralba, Torresani, Yan, and Malik]{Grauman_2022_CVPR}
Kristen Grauman, Andrew Westbury, Eugene Byrne, Zachary Chavis, Antonino
  Furnari, Rohit Girdhar, Jackson Hamburger, Hao Jiang, Miao Liu, Xingyu Liu,
  Miguel Martin, Tushar Nagarajan, Ilija Radosavovic, Santhosh~Kumar
  Ramakrishnan, Fiona Ryan, Jayant Sharma, Michael Wray, Mengmeng Xu,
  Eric~Zhongcong Xu, Chen Zhao, Siddhant Bansal, Dhruv Batra, Vincent
  Cartillier, Sean Crane, Tien Do, Morrie Doulaty, Akshay Erapalli, Christoph
  Feichtenhofer, Adriano Fragomeni, Qichen Fu, Abrham Gebreselasie, Cristina
  Gonz\'alez, James Hillis, Xuhua Huang, Yifei Huang, Wenqi Jia, Weslie Khoo,
  J\'achym Kol\'a\v{r}, Satwik Kottur, Anurag Kumar, Federico Landini, Chao Li,
  Yanghao Li, Zhenqiang Li, Karttikeya Mangalam, Raghava Modhugu, Jonathan
  Munro, Tullie Murrell, Takumi Nishiyasu, Will Price, Paola Ruiz, Merey
  Ramazanova, Leda Sari, Kiran Somasundaram, Audrey Southerland, Yusuke Sugano,
  Ruijie Tao, Minh Vo, Yuchen Wang, Xindi Wu, Takuma Yagi, Ziwei Zhao, Yunyi
  Zhu, Pablo Arbel\'aez, David Crandall, Dima Damen, Giovanni~Maria Farinella,
  Christian Fuegen, Bernard Ghanem, Vamsi~Krishna Ithapu, C.~V. Jawahar,
  Hanbyul Joo, Kris Kitani, Haizhou Li, Richard Newcombe, Aude Oliva, Hyun~Soo
  Park, James~M. Rehg, Yoichi Sato, Jianbo Shi, Mike~Zheng Shou, Antonio
  Torralba, Lorenzo Torresani, Mingfei Yan, and Jitendra Malik.
\newblock Ego4d: Around the world in 3,000 hours of egocentric video.
\newblock In \emph{CVPR}, June 2022.

\bibitem[Gu et~al.(2022)Gu, Lin, Kuo, and Cui]{gu2022openvocabulary}
Xiuye Gu, Tsung-Yi Lin, Weicheng Kuo, and Yin Cui.
\newblock Open-vocabulary object detection via vision and language knowledge
  distillation.
\newblock In \emph{ICLR}, 2022.

\bibitem[Gupta et~al.(2019)Gupta, Dollar, and Girshick]{lvis}
Agrim Gupta, Piotr Dollar, and Ross Girshick.
\newblock Lvis: A dataset for large vocabulary instance segmentation.
\newblock In \emph{CVPR}, 2019.

\bibitem[Hayat et~al.(2020)Hayat, Hayat, Rahman, Khan, Zamir, and
  Khan]{hayat2020synthesizing}
Nasir Hayat, Munawar Hayat, Shafin Rahman, Salman Khan, Syed~Waqas Zamir, and
  Fahad~Shahbaz Khan.
\newblock Synthesizing the unseen for zero-shot object detection.
\newblock In \emph{ACCV}, 2020.

\bibitem[He et~al.(2016)He, Zhang, Ren, and Sun]{he2016deep}
Kaiming He, Xiangyu Zhang, Shaoqing Ren, and Jian Sun.
\newblock Deep residual learning for image recognition.
\newblock In \emph{CVPR}, 2016.

\bibitem[He et~al.(2017)He, Gkioxari, Doll{\'a}r, and Girshick]{he2017mask}
Kaiming He, Georgia Gkioxari, Piotr Doll{\'a}r, and Ross Girshick.
\newblock Mask r-cnn.
\newblock In \emph{ICCV}, 2017.

\bibitem[He \& Peng(2017)He and Peng]{he2017fine}
Xiangteng He and Yuxin Peng.
\newblock Fine-grained image classification via combining vision and language.
\newblock In \emph{CVPR}, 2017.

\bibitem[Hu et~al.(2022)Hu, Gan, Wang, Yang, Liu, Lu, and Wang]{Hu_2022_CVPR}
Xiaowei Hu, Zhe Gan, Jianfeng Wang, Zhengyuan Yang, Zicheng Liu, Yumao Lu, and
  Lijuan Wang.
\newblock Scaling up vision-language pre-training for image captioning.
\newblock In \emph{CVPR}, pp.\  17980--17989, June 2022.

\bibitem[Jayaraman \& Grauman(2014)Jayaraman and Grauman]{jayaraman2014zero}
Dinesh Jayaraman and Kristen Grauman.
\newblock Zero shot recognition with unreliable attributes.
\newblock In \emph{NeurIPS}, 2014.

\bibitem[Jia et~al.(2021)Jia, Yang, Xia, Chen, Parekh, Pham, Le, Sung, Li, and
  Duerig]{align}
Chao Jia, Yinfei Yang, Ye~Xia, Yi-Ting Chen, Zarana Parekh, Hieu Pham, Quoc~V
  Le, Yunhsuan Sung, Zhen Li, and Tom Duerig.
\newblock Scaling up visual and vision-language representation learning with
  noisy text supervision.
\newblock In \emph{ICML}, 2021.

\bibitem[Joulin et~al.(2016)Joulin, Maaten, Jabri, and
  Vasilache]{joulin2016learning}
Armand Joulin, Laurens van~der Maaten, Allan Jabri, and Nicolas Vasilache.
\newblock Learning visual features from large weakly supervised data.
\newblock In \emph{ECCV}, 2016.

\bibitem[Kim et~al.(2022)Kim, Lin, Angelova, Kweon, and Kuo]{oln}
Dahun Kim, Tsung-Yi Lin, Anelia Angelova, In~So Kweon, and Weicheng Kuo.
\newblock Learning open-world object proposals without learning to classify.
\newblock \emph{IEEE Robotics and Automation Letters}, 2022.

\bibitem[Kuo et~al.(2015)Kuo, Hariharan, and Malik]{Kuo_2015_ICCV}
Weicheng Kuo, Bharath Hariharan, and Jitendra Malik.
\newblock Deepbox: Learning objectness with convolutional networks.
\newblock In \emph{ICCV}, 2015.

\bibitem[Li et~al.(2021)Li, Selvaraju, Gotmare, Joty, Xiong, and
  Hoi]{li2021align}
Junnan Li, Ramprasaath Selvaraju, Akhilesh Gotmare, Shafiq Joty, Caiming Xiong,
  and Steven Chu~Hong Hoi.
\newblock Align before fuse: Vision and language representation learning with
  momentum distillation.
\newblock \emph{NeurIPS}, 34:\penalty0 9694--9705, 2021.

\bibitem[Li et~al.(2022{\natexlab{a}})Li, Li, Xiong, and Hoi]{li2022blip}
Junnan Li, Dongxu Li, Caiming Xiong, and Steven Hoi.
\newblock Blip: Bootstrapping language-image pre-training for unified
  vision-language understanding and generation.
\newblock \emph{arXiv preprint arXiv:2201.12086}, 2022{\natexlab{a}}.

\bibitem[Li et~al.(2022{\natexlab{b}})Li, Zhang, Zhang, Yang, Li, Zhong, Wang,
  Yuan, Zhang, Hwang, Chang, and Gao]{li2021grounded}
Liunian~Harold Li, Pengchuan Zhang, Haotian Zhang, Jianwei Yang, Chunyuan Li,
  Yiwu Zhong, Lijuan Wang, Lu~Yuan, Lei Zhang, Jenq-Neng Hwang, Kai-Wei Chang,
  and Jianfeng Gao.
\newblock Grounded language-image pre-training.
\newblock In \emph{CVPR}, 2022{\natexlab{b}}.

\bibitem[Li et~al.(2022{\natexlab{c}})Li, Mao, Girshick, and
  He]{li2022exploring}
Yanghao Li, Hanzi Mao, Ross Girshick, and Kaiming He.
\newblock Exploring plain vision transformer backbones for object detection.
\newblock In \emph{ECCV}, 2022{\natexlab{c}}.

\bibitem[Lin et~al.(2014)Lin, Maire, Belongie, Hays, Perona, Ramanan,
  Doll{\'a}r, and Zitnick]{coco}
Tsung-Yi Lin, Michael Maire, Serge Belongie, James Hays, Pietro Perona, Deva
  Ramanan, Piotr Doll{\'a}r, and C~Lawrence Zitnick.
\newblock Microsoft coco: Common objects in context.
\newblock In \emph{ECCV}, 2014.

\bibitem[Lin et~al.(2017)Lin, Doll{\'a}r, Girshick, He, Hariharan, and
  Belongie]{fpn}
Tsung-Yi Lin, Piotr Doll{\'a}r, Ross Girshick, Kaiming He, Bharath Hariharan,
  and Serge Belongie.
\newblock Feature pyramid networks for object detection.
\newblock In \emph{CVPR}, 2017.

\bibitem[Minderer et~al.(2022)Minderer, Gritsenko, Stone, Neumann, Weissenborn,
  Dosovitskiy, Mahendran, Arnab, Dehghani, Shen, Wang, Zhai, Kipf, and
  Houlsby]{minderer2022simple}
Matthias Minderer, Alexey Gritsenko, Austin Stone, Maxim Neumann, Dirk
  Weissenborn, Alexey Dosovitskiy, Aravindh Mahendran, Anurag Arnab, Mostafa
  Dehghani, Zhuoran Shen, Xiao Wang, Xiaohua Zhai, Thomas Kipf, and Neil
  Houlsby.
\newblock Simple open-vocabulary object detection with vision transformers.
\newblock In \emph{ECCV}, 2022.

\bibitem[Norouzi et~al.(2014)Norouzi, Mikolov, Bengio, Singer, Shlens, Frome,
  Corrado, and Dean]{norouzi2013zero}
Mohammad Norouzi, Tomas Mikolov, Samy Bengio, Yoram Singer, Jonathon Shlens,
  Andrea Frome, Greg~S Corrado, and Jeffrey Dean.
\newblock Zero-shot learning by convex combination of semantic embeddings.
\newblock 2014.

\bibitem[Pham et~al.(2021)Pham, Dai, Ghiasi, Liu, Yu, Luong, Tan, and
  Le]{basic}
Hieu Pham, Zihang Dai, Golnaz Ghiasi, Hanxiao Liu, Adams~Wei Yu, Minh{-}Thang
  Luong, Mingxing Tan, and Quoc~V. Le.
\newblock Combined scaling for zero-shot transfer learning.
\newblock \emph{CoRR}, abs/2111.10050, 2021.
\newblock URL \url{https://arxiv.org/abs/2111.10050}.

\bibitem[Radford et~al.(2021)Radford, Kim, Hallacy, Ramesh, Goh, Agarwal,
  Sastry, Askell, Mishkin, Clark, Krueger, and Sutskever]{radford2021clip}
Alec Radford, Jong~Wook Kim, Chris Hallacy, Aditya Ramesh, Gabriel Goh,
  Sandhini Agarwal, Girish Sastry, Amanda Askell, Pamela Mishkin, Jack Clark,
  Gretchen Krueger, and Ilya Sutskever.
\newblock Learning transferable visual models from natural language
  supervision.
\newblock In \emph{ICML}, 2021.

\bibitem[Rahman et~al.(2020)Rahman, Khan, and Barnes]{rahman2020improved}
Shafin Rahman, Salman Khan, and Nick Barnes.
\newblock Improved visual-semantic alignment for zero-shot object detection.
\newblock In \emph{AAAI}, 2020.

\bibitem[Redmon et~al.(2016)Redmon, Divvala, Girshick, and
  Farhadi]{redmon2016you}
Joseph Redmon, Santosh Divvala, Ross Girshick, and Ali Farhadi.
\newblock You only look once: Unified, real-time object detection.
\newblock In \emph{CVPR}, 2016.

\bibitem[Ren et~al.(2015)Ren, He, Girshick, and Sun]{ren2015faster}
Shaoqing Ren, Kaiming He, Ross~B Girshick, and Jian Sun.
\newblock Faster r-cnn: Towards real-time object detection with region proposal
  networks.
\newblock In \emph{NeurIPS}, 2015.

\bibitem[Rohrbach et~al.(2011)Rohrbach, Stark, and
  Schiele]{rohrbach2011evaluating}
Marcus Rohrbach, Michael Stark, and Bernt Schiele.
\newblock Evaluating knowledge transfer and zero-shot learning in a large-scale
  setting.
\newblock In \emph{CVPR}, 2011.

\bibitem[Sariyildiz et~al.(2020)Sariyildiz, Perez, and
  Larlus]{sariyildiz2020learning}
Mert~Bulent Sariyildiz, Julien Perez, and Diane Larlus.
\newblock Learning visual representations with caption annotations.
\newblock In \emph{ECCV}, 2020.

\bibitem[Shao et~al.(2019)Shao, Li, Zhang, Peng, Yu, Zhang, Li, and
  Sun]{objects365}
Shuai Shao, Zeming Li, Tianyuan Zhang, Chao Peng, Gang Yu, Xiangyu Zhang, Jing
  Li, and Jian Sun.
\newblock Objects365: A large-scale, high-quality dataset for object detection.
\newblock In \emph{ICCV}, 2019.

\bibitem[Sharma et~al.(2018)Sharma, Ding, Goodman, and Soricut]{ccdataset}
Piyush Sharma, Nan Ding, Sebastian Goodman, and Radu Soricut.
\newblock Conceptual captions: A cleaned, hypernymed, image alt-text dataset
  for automatic image captioning.
\newblock In \emph{ACL}, 2018.

\bibitem[Singh et~al.(2022)Singh, Hu, Goswami, Couairon, Galuba, Rohrbach, and
  Kiela]{singh2022flava}
Amanpreet Singh, Ronghang Hu, Vedanuj Goswami, Guillaume Couairon, Wojciech
  Galuba, Marcus Rohrbach, and Douwe Kiela.
\newblock Flava: A foundational language and vision alignment model.
\newblock In \emph{CVPR}, pp.\  15638--15650, 2022.

\bibitem[Vasconcelos et~al.(2022)Vasconcelos, Birodkar, and
  Dumoulin]{vasconcelos2022proper}
Cristina Vasconcelos, Vighnesh Birodkar, and Vincent Dumoulin.
\newblock Proper reuse of image classification features improves object
  detection.
\newblock In \emph{CVPR}, 2022.

\bibitem[Wang et~al.(2009)Wang, Markert, Everingham, et~al.]{wang2009learning}
Josiah Wang, Katja Markert, Mark Everingham, et~al.
\newblock Learning models for object recognition from natural language
  descriptions.
\newblock In \emph{BMVC}, 2009.

\bibitem[Wang et~al.(2020)Wang, Mahajan, and Ramanathan]{wang2020leads}
Rui Wang, Dhruv Mahajan, and Vignesh Ramanathan.
\newblock What leads to generalization of object proposals?
\newblock In \emph{ECCV}, 2020.

\bibitem[Wang et~al.(2021)Wang, Yu, Yu, Dai, Tsvetkov, and Cao]{wang2021simvlm}
Zirui Wang, Jiahui Yu, Adams~Wei Yu, Zihang Dai, Yulia Tsvetkov, and Yuan Cao.
\newblock Simvlm: Simple visual language model pretraining with weak
  supervision.
\newblock \emph{arXiv preprint arXiv:2108.10904}, 2021.

\bibitem[Wei et~al.(2021)Wei, Gao, Wu, Hu, and Lin]{SoCo}
Fangyun Wei, Yue Gao, Zhirong Wu, Han Hu, and Stephen Lin.
\newblock Aligning pretraining for detection via object-level contrastive
  learning.
\newblock In \emph{NeurIPS}, 2021.

\bibitem[Wu et~al.(2019)Wu, Kirillov, Massa, Lo, and
  Girshick]{wu2019detectron2}
Yuxin Wu, Alexander Kirillov, Francisco Massa, Wan-Yen Lo, and Ross Girshick.
\newblock Detectron2.
\newblock \url{https://github.com/facebookresearch/detectron2}, 2019.

\bibitem[Ye et~al.(2019)Ye, Zhang, Kovashka, Li, Qin, and Berent]{Cap2Det}
Keren Ye, Mingda Zhang, Adriana Kovashka, Wei Li, Danfeng Qin, and Jesse
  Berent.
\newblock Cap2det: Learning to amplify weak caption supervision for object
  detection.
\newblock In \emph{ICCV}, 2019.

\bibitem[Zareian et~al.(2021)Zareian, Rosa, Hu, and Chang]{Zareian_2021_CVPR}
Alireza Zareian, Kevin~Dela Rosa, Derek~Hao Hu, and Shih-Fu Chang.
\newblock Open-vocabulary object detection using captions.
\newblock In \emph{CVPR}, 2021.

\bibitem[Zhai et~al.(2022)Zhai, Wang, Mustafa, Steiner, Keysers, Kolesnikov,
  and Beyer]{zhai2021lit}
Xiaohua Zhai, Xiao Wang, Basil Mustafa, Andreas Steiner, Daniel Keysers,
  Alexander Kolesnikov, and Lucas Beyer.
\newblock Lit: Zero-shot transfer with locked-image text tuning.
\newblock In \emph{CVPR}, 2022.

\bibitem[Zhao et~al.(2022)Zhao, Zhang, Schulter, Zhao, Stathopoulos,
  Chandraker, Metaxas, et~al.]{zhao2022exploiting}
Shiyu Zhao, Zhixing Zhang, Samuel Schulter, Long Zhao, Anastasis Stathopoulos,
  Manmohan Chandraker, Dimitris Metaxas, et~al.
\newblock Exploiting unlabeled data with vision and language models for object
  detection.
\newblock In \emph{ECCV}, 2022.

\bibitem[Zheng et~al.(2020)Zheng, Huang, Han, Huang, and
  Cui]{zheng2020background}
Ye~Zheng, Ruoran Huang, Chuanqi Han, Xi~Huang, and Li~Cui.
\newblock Background learnable cascade for zero-shot object detection.
\newblock In \emph{ACCV}, 2020.

\bibitem[Zhong et~al.(2021)Zhong, Shi, Yang, Xu, and Li]{zhong2021learning}
Yiwu Zhong, Jing Shi, Jianwei Yang, Chenliang Xu, and Yin Li.
\newblock Learning to generate scene graph from natural language supervision.
\newblock In \emph{ICCV}, 2021.

\bibitem[Zhong et~al.(2022)Zhong, Yang, Zhang, Li, Codella, Li, Zhou, Dai,
  Yuan, Li, and Gao]{zhong2021regionclip}
Yiwu Zhong, Jianwei Yang, Pengchuan Zhang, Chunyuan Li, Noel Codella,
  Liunian~Harold Li, Luowei Zhou, Xiyang Dai, Lu~Yuan, Yin Li, and Jianfeng
  Gao.
\newblock Regionclip: Region-based language-image pretraining.
\newblock In \emph{CVPR}, 2022.

\bibitem[Zhou et~al.(2022{\natexlab{a}})Zhou, Loy, and Dai]{zhou2022maskclip}
Chong Zhou, Chen~Change Loy, and Bo~Dai.
\newblock Extract free dense labels from clip.
\newblock In \emph{ECCV}, 2022{\natexlab{a}}.

\bibitem[Zhou et~al.(2022{\natexlab{b}})Zhou, Yang, Loy, and
  Liu]{zhou2022cocoop}
Kaiyang Zhou, Jingkang Yang, Chen~Change Loy, and Ziwei Liu.
\newblock Conditional prompt learning for vision-language models.
\newblock In \emph{CVPR}, 2022{\natexlab{b}}.

\bibitem[Zhou et~al.(2022{\natexlab{c}})Zhou, Girdhar, Joulin,
  Kr{\"a}henb{\"u}hl, and Misra]{zhou2022detecting}
Xingyi Zhou, Rohit Girdhar, Armand Joulin, Philipp Kr{\"a}henb{\"u}hl, and
  Ishan Misra.
\newblock Detecting twenty-thousand classes using image-level supervision.
\newblock In \emph{ECCV}, 2022{\natexlab{c}}.

\bibitem[Zhu et~al.(2020)Zhu, Wang, and Saligrama]{zhu2020don}
Pengkai Zhu, Hanxiao Wang, and Venkatesh Saligrama.
\newblock Don't even look once: Synthesizing features for zero-shot detection.
\newblock In \emph{CVPR}, 2020.

\end{thebibliography}
\bibliographystyle{iclr2023_conference}

\newpage
\appendix
\section*{Appendix}

\section{Ablation}
\label{sec:ablations}
\subsection{Finetuning versus Frozen Backbone} 
\label{sec:backbone-finetuning}
We explore the pros and cons of backbone finetuning compared to the frozen backbone. We observe that finetuning the backbone with the same training recipe diverges, so we apply gradient clipping (max norm = $1.0$) and reduce the backbone learning rate significantly. Table~\ref{tab:backbone-finetuning} shows that although finetuning can benefit the base categories, it slightly compromises the novel category with higher memory/compute footprint.

\begin{table}[h]
\caption{\textbf{Finetuning vs frozen backbone.} Finetuning does not benefit the novel categories (AP$_r$) but improves the base categories (AP$_c$, AP$_f$).}
\centering
\begin{tabular}{l|l>{\color{gray}}c>{\color{gray}}c>{\color{gray}}c}
 \toprule
 Backbone LR & AP$_r$ & AP$_c$ & AP$_f$ & AP \\
 \midrule
 1e-3 & 18.1 & 25.7 & 30.2 & 26.2 \\
 1e-4 & 18.1 & 24.9 & 28.8 & 25.3 \\
 0.0 & 18.6 (\textcolor{teal}{+0.5}) & 24.0 & 26.9 & 24.2 \\
\bottomrule
\end{tabular}
\label{tab:backbone-finetuning}
\end{table}

\subsection{Score Fusion} 
\label{sec:score-fuse}
\begin{figure*}[h]
    \centering
	\includegraphics[width=0.48\linewidth]{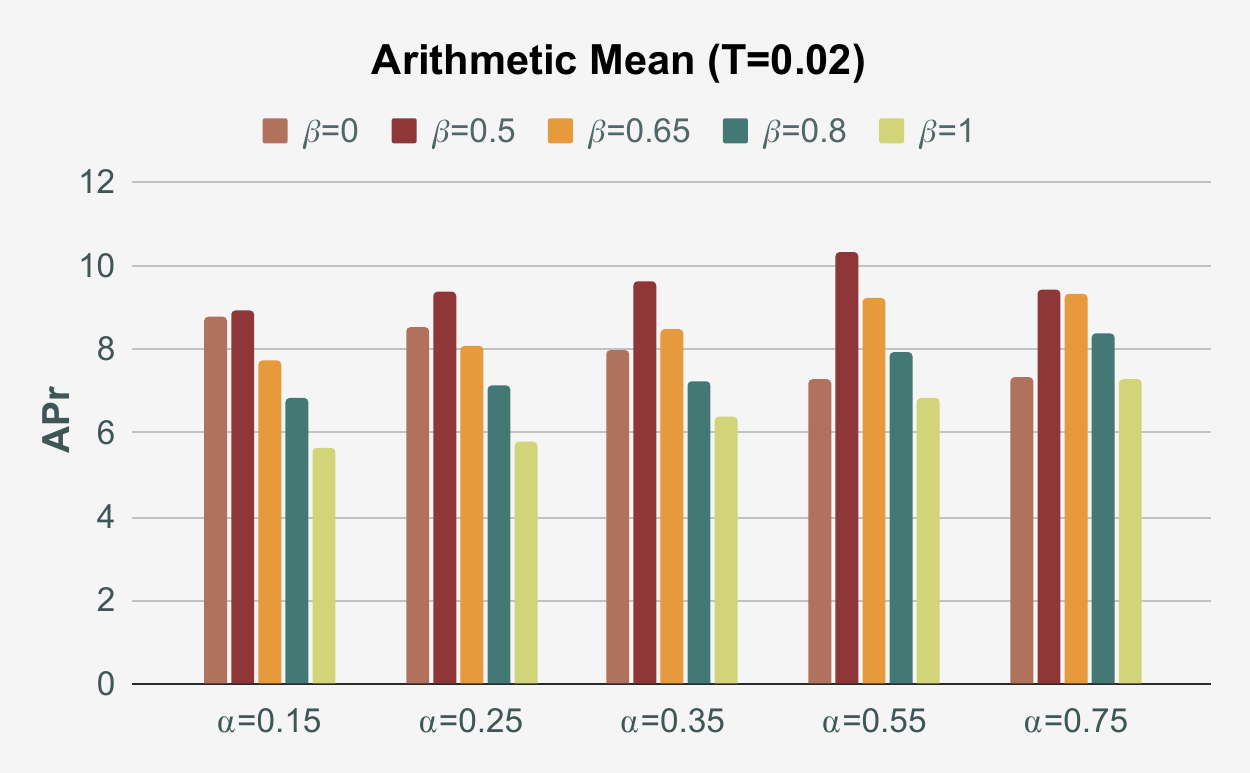}
	\includegraphics[width=0.48\linewidth]{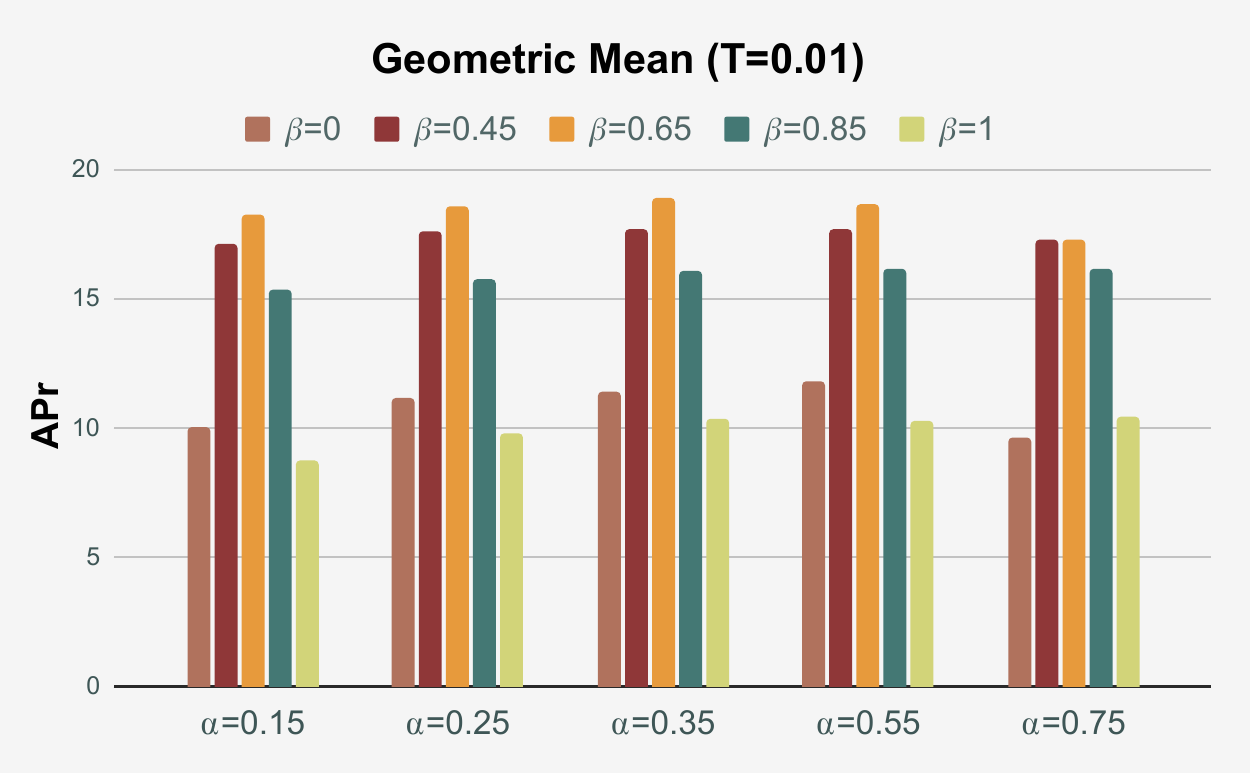}
	\caption{\textbf{Hyper-parameter sweep on score fusion parameters}. We observe that geometric means (right) are significantly better than arithmetic means (left). All results are based on a trained F-VLM R50 model.}
	\label{fig:sweep-score-fuse}
\end{figure*}

We explore the use of arithmetic vs geometric means to fuse the VLM and detection scores in~\eqref{eqn:combine-score}. Table~\ref{tab:score-fuse} shows that using geometric mean is significantly better by more than 8 points. In Figure~\ref{fig:sweep-score-fuse}, we perform a dense grid sweep over $\alpha, \beta$ and confirm the 8-point gap between geometric and arithmetic means still holds.

We perform a more in-depth study of score fusion parameters in Table~\ref{tab:score-ablations}. From the table, we see that $\beta$ is the main tunable parameter of our model, and the performance is relatively robust to $\alpha$. For most practical use cases, we recommend setting $T=0.01$. The temperature $\tau$ in Equation~\ref{eqn:detect-logit} is learned automatically and needs no tuning.

\begin{table}[h]
\caption{\textbf{Score Fusion.} We study different score fusion mechanisms of \OURS{}. We report AP$_r$ and AP on LVIS. Geometric mean is significantly more effective than arithmetic mean. All results are average over 5 independent runs using R50 backbone.}
\centering
\begin{tabular}{lccc|l>{\color{gray}}c}
 \toprule
 Fusion & $\beta$ & $\alpha$ & $T$ & AP$_r$ & AP \\
 \midrule
 Arithmetic & 0.65 & 0.35 & 0.01 & 9.1 & 16.4 \\
 Arithmetic & 0.65 & 0.35 & 0.02 & 9.3 & 19.8 \\
 Arithmetic & 0.5 & 0.75 & 0.02 & 10.3 & 15.9 \\
 Geometric & 0.65 & 0.35 & 0.01 & 18.6 (\textcolor{teal}{+8.3}) & 24.2 \\
\bottomrule
\end{tabular}
\label{tab:score-fuse}
\end{table}

\begin{table*}[h]
    \caption{\textbf{Score Fusion Parameters}. We study different geometric mean fusion parameters by comparing their AP$_r$ on LVIS. All results are based on a trained \OURS{} R50 model. Default settings are in \colorbox{baselinecolor}{gray}.}
    \centering
    \vspace{-1mm}
    \begin{subtable}{0.33 \linewidth}
    \caption{\textbf{VLM-score weights for novel classes.} We fix $\alpha=0.35$ and $T=0.01$.}
    \label{tab:score-novel}
    \centering
    \scalebox{0.92}{
    \begin{tabular}{l|c>{\color{gray}}c}
     \toprule
     $\beta$ & AP$_r$ & AP \\
     \midrule
     0.0 & 11.3 & 22.9 \\
     0.45 & 17.8 & 24.2 \\
     0.65 & \baseline{\textbf{18.9}} & \baseline{\textbf{24.2}} \\
     0.85 & 16.1 & 22.5 \\
     1.00 & 10.3 & 17.8 \\
    \bottomrule
    \end{tabular}
    }
    \end{subtable}
    \hfill
    \begin{subtable}[h]{0.33 \linewidth}
    \centering
    \caption{\textbf{VLM-score weights for base classes.} We fix $\beta=0.65$ and $T=0.01$.}
    \label{tab:score-base}
    \scalebox{0.92}{
    \begin{tabular}{l|c>{\color{gray}}c}
     \toprule
     $\alpha$ & AP$_r$ & AP \\
     \midrule
     0.15 & 18.2 & 24.3 \\
     0.25 & 18.8 & 24.5 \\
     0.35 & \baseline{\textbf{18.9}} & \baseline{\textbf{24.2}} \\
     0.55 & 18.7 & 22.4 \\
     0.75 & 17.3 & 17.8 \\
    \bottomrule
    \end{tabular}
    }
    \end{subtable} 
    \hfill
    \begin{subtable}[h]{0.27 \linewidth}
    \centering
    \caption{\textbf{Temperature for VLM logits.} We fix $\beta=0.65$ and $\alpha=0.35$}
    \label{tab:temperature}
    \scalebox{0.92}{
    \begin{tabular}{l|c>{\color{gray}}c}
     \toprule
     $T$ & AP$_r$ & AP \\
     \midrule
     0.0025 & 14.9 & 18.7 \\
     0.005 & 17.3 & 22.2 \\
     0.01 & \baseline{\textbf{18.9}} & \baseline{\textbf{24.2}} \\
     0.02 & 17.8 & 24.8 \\
     0.04 & 13.9 & 24.4 \\
    \bottomrule
    \end{tabular}
    }
    \end{subtable}
    \label{tab:score-ablations}
\end{table*}

\subsection{Feature Pyramid Capacity}
\label{sec:large-fpn}
We explore the effects of increasing the feature pyramid~\citep{fpn} capacity to enhance the representation learned upon the frozen backbone features. To increase the FPN capacity, we simply repeat the lateral and top-down connections of FPN $N$ times before applying the post-hoc convolution. We insert a ReLU and BatchNorm layer in each lateral connection, and add a skip connection from the backbone feature maps to every level (\ie $1, 2, ..., N$). The post-hoc convolution and BatchNorm layers are kept the same. All runs are repeated 5 times and trained for 46.1k steps following the same protocol as the LVIS benchmark of the manuscript.

Table~\ref{tab:large-fpn} shows that increased feature pyramid capacity improves the base categories significantly (AP$_c$, AP$_f$) without compromising the novel categories (AP$_r$). Although it is common to improve the base categories at the cost of novel categories, enlarged feature pyramid leads to improvements on all categories including a slight improvement of 0.1 on AP$_r$.

\begin{table}[h]
\caption{\textbf{Feature Pyramid Capacity.} We observe that larger feature pyramid improves the base categories (AP$_c$, AP$_f$) without compromising the novel categories (AP$_r$).}
\centering
\begin{tabular}{llllll}
 \toprule
 Backbone & Feature Pyramid & AP$_r$ & AP$_c$ & AP$_f$ & AP \\
 \midrule
 R50x64 & FPN~\citep{fpn} & 32.8 & 35.4 & 35.4 & 34.9 \\
 R50x64 & FPN ($N=12$) & 32.9 (\textcolor{teal}{+0.1}) & 37.5 (\textcolor{teal}{+1.9}) & 38.3 (\textcolor{teal}{+2.9}) & 37.0 (\textcolor{teal}{+2.1}) \\
\bottomrule
\end{tabular}
\label{tab:large-fpn}
\end{table}

\subsection{Background Weight} 
\label{sec:background-loss}
We study the effects of background weight~\citep{Zareian_2021_CVPR,zhong2021regionclip} on \OURS{}. Consistent with the findings in~\citep{zhong2021regionclip}, we found that a background weight of 0.9 is slightly better than the default 1.0 in Table~\ref{tab:bg-weight}. Therefore, we use background weight 0.9 as default. All results are average over 3 independent runs.

\begin{table}[h]
\caption{\textbf{Background Weight.} We study the effects of background weight of \OURS{} on LVIS.}
\centering
\begin{tabular}{lc|l}
 \toprule
 Backbone & Background Weight & AP$_r$ \\
 \midrule
 R50 & 1.0 & 18.3 \\
 R50 & 0.9 & 18.4 (\textcolor{teal}{+0.1}) \\
 \midrule
 R50x64 & 1.0 & 32.4 \\
 R50x64 & 0.9 & 32.9 (\textcolor{teal}{+0.5}) \\
\bottomrule
\end{tabular}
\label{tab:bg-weight}
\end{table}

\section{Computation-friendly Training}
\label{sec:standard-recipe}
To facilitate comparison with the broader research community, we validate the efficacy of \OURS{} in more computation-friendly 1$\times$ (12 epochs) and 3$\times$ (36 epochs) settings~\citep{wu2019detectron2} by using smaller batch size and no large-scale-jittering (LSJ) augmentation~\citep{ghiasi2021simple}. The results are listed in Table~\ref{tab:standard-recipe}.

\begin{table}[h]
\caption{\textbf{Benchmark on computation-friendly training recipes.} By leveraging frozen backbone, \OURS{} is robust to shorter schedule and smaller batch size. All results are reported as the average over 5 runs. Our default settings are in \colorbox{baselinecolor}{gray}.}
\centering
\begin{tabular}{l|ccc|c}
 \toprule
 Backbone & LSJ & \# Epochs & Batch Size & AP$_r$ \\
 \midrule
 R50 & & 12 (1$\times$)  & 16 & 18.1 \\
 R50 & & 36 (3$\times$) & 64 & 18.5 \\
 \baseline{R50} & \baseline{$\checkmark$}& \baseline{100} & \baseline{256} & \baseline{18.6} \\
 \midrule
 R50x64 & & 12 (1$\times$)  & 16 & 31.9 \\
 R50x64 & & 36 (3$\times$) & 64 & 32.6 \\
 \baseline{R50x64} & \baseline{$\checkmark$} & \baseline{100} & \baseline{256} & \baseline{32.8} \\
\bottomrule
\end{tabular}
\label{tab:standard-recipe}
\end{table}

We observe that \OURS{} is robust to the number of training epochs, batch size, with/without LSJ~\citep{ghiasi2021simple} for both the smallest and largest backbones. This stands in contrast to the sensitivity of fully supervised learning to these hyper-parameters, and is consistent with our findings in Table~\ref{table:compute} that frozen backbone contributes to the training efficiency and stability.

\section{Exploring the Structure of Frozen Features}
\label{sec:vis-clustering}
\begin{figure*}[t]
    \centering
	\includegraphics[width=1.00\linewidth]{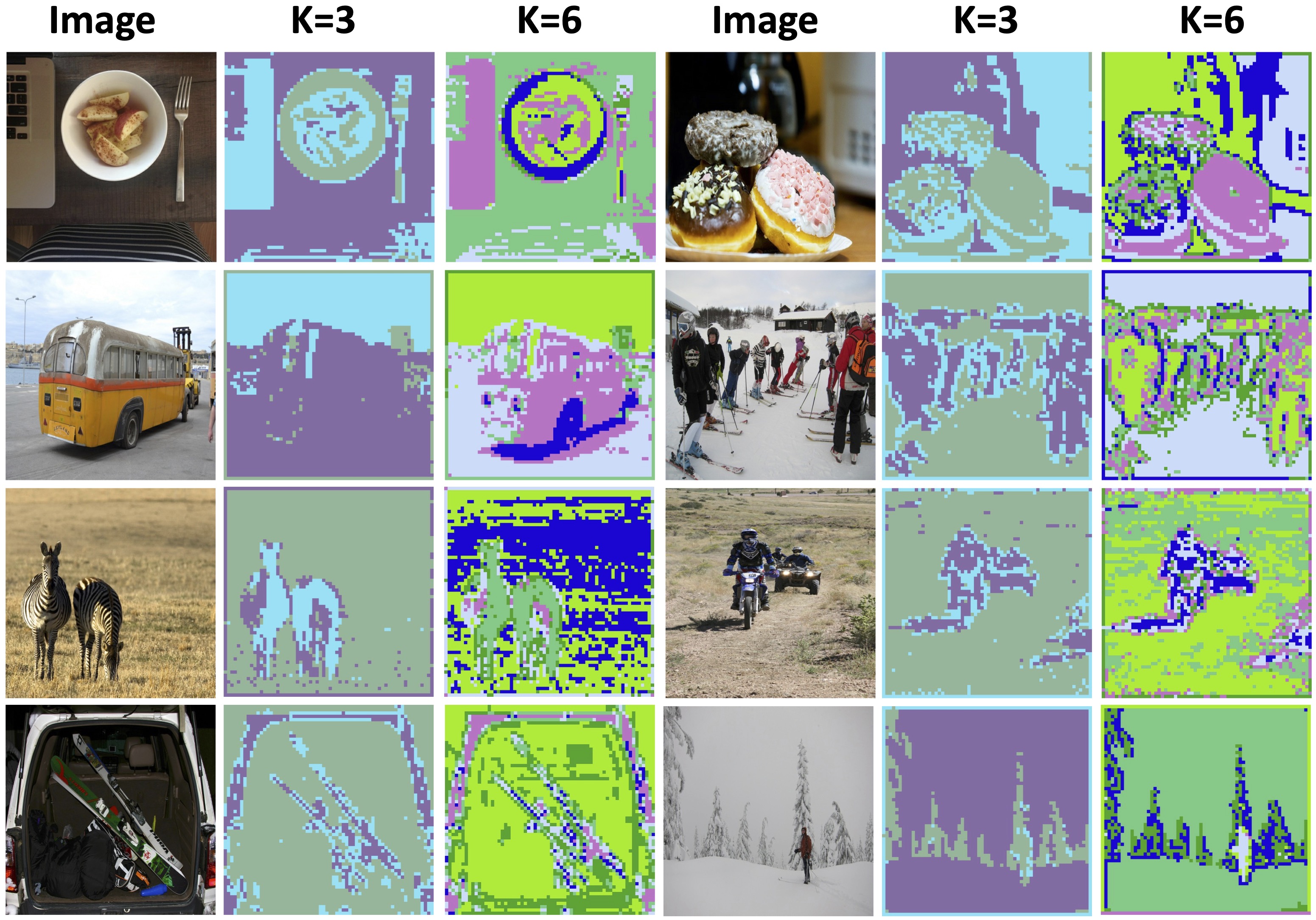}
	\caption{\textbf{Understanding the frozen VLM feature clusters}. Salient objects and object parts emerge naturally from the clustering of frozen VLM features.}
	\label{fig:vis-clustering}
\end{figure*}

To understand the effectiveness of F-VLM, we perform k-means clustering to probe the structures present in the frozen VLM features (\eg CLIP). We use CLIP R50x4 backbone and LVIS dataset for visualization. Only the last layer output features are used for clustering, because these features can be used for zero-shot region classification at the same time. Figure~\ref{fig:vis-clustering} demonstrates that the features form nice clusters around salient objects of the scenes (\eg, skis, motorbikes, people), and naturally separate object parts (\eg, donut toppings, bus wheels) without explicit supervision. We believe these emergent properties of frozen VLM are promising avenues to push open-vocabulary detection beyond the domains of existing detection datasets.

\section{Analysis of Transfer Detection Benchmark}
\label{sec:vocab-overlap}
Many existing works benchmark transfer detection across common detection datasets~\citep{gu2022openvocabulary,du2022learning,li2021grounded,minderer2022simple}. In particular, the LVIS to COCO and Objects365 transfer detection is a recent benchmark proposed by~\cite{gu2022openvocabulary}. We analyze and report the vocabulary overlap percentage between COCO/Objects365 and LVIS base categories in Table~\ref{tab:vocab-overlap}.

\begin{table}[h]
\caption{\textbf{Transfer Detection Vocabulary Overlap.} We observe substantial overlap between COCO and Objects365 vocabulary and LVIS base categories.}
\centering
\begin{tabular}{l|cc}
 \toprule
 Method & COCO & Objects365 \\
 \midrule
 Name matching & 85\% & 56\% \\
 Name matching + near duplicate removal & 91\% & 63\% \\
\bottomrule
\end{tabular}
\label{tab:vocab-overlap}
\end{table}

Simple name matching shows clear overlap, while the removal of near duplicates (e.g. synonyms and non-alphabetic character removal) reveals even more. In addition, we notice that COCO has more overlap than Objects365 due to its smaller vocabulary. These results show the limitation of existing transfer detection setup, and we encourage the community to move towards larger transfer detection benchmarks with less vocabulary overlap.

\section{Memory Use}
\label{sec:memory-use}
The memory consumption of F-VLM is almost the same as Mask R-CNN, with all class-specific heads changed to class-agnostic. F-VLM uses the same batch size as~\cite{gu2022openvocabulary} in the default settings, although we found  it works well even with much smaller batch size and epoch length (see Appendix~\ref{sec:standard-recipe}). 

Moreover, F-VLM has significant memory saving potential compared to existing approaches that fine-tune the backbone, especially with large backbones. At training time, F-VLM does not need to store forward-pass activations, gradients or gradient moments, and the memory use of the backbone is just the backbone weights plus a small amount of current activations. This makes F-VLM highly memory efficient especially with large backbones. In practice, the actual memory use depends on the low-level implementation of each deep learning library. In pytorch, for example, ``torch.no\_grad()'' context manager~\footnote{\href{https://pytorch.org/docs/stable/generated/torch.no_grad.html}{https://pytorch.org/docs/stable/generated/torch.no\_grad.html}} can enable such behavior. 

\section{Can we use other VLMs?}
\label{sec:other-vlm}
In this work, we adopt the widely used CLIP~\citep{radford2021clip} to develop a simple open-vocabulary detection recipe based on frozen backbones. Moving forward, we believe it would be very interesting to explore different pretrained VLMs, which may involve substantial modifications to \OURS{}. For example, ViT-based pretrained VLMs~\citep{radford2021clip,li2021align,li2022blip,zhai2021lit} have become popular recently. These VLMs require single-scale ViT-based detector such as ViTDet~\citep{li2022exploring} to adapt them for open-vocabulary detection. To use VLMs like ALBEF~\citep{li2021align} and BLIP~\citep{li2022blip}, it is important to consider how to efficiently compute all-pair region-text similarities with multimodal encoders (as opposed to dual encoders). Furthermore, it is an open question how to use captioning VLMs~\citep{wang2021simvlm,Hu_2022_CVPR} or masked multimodal VLMs~\citep{singh2022flava} for open-vocabulary detection. We believe these are all interesting directions for the community to explore.   

\section{Implementation Details}
\label{sec:appendix-impl}
Table~\ref{tab:hyperparams} summarizes the hyper-parameters we use for LVIS and COCO experiments. On LVIS, we adopt the same hyper-parameters as~\citet{gu2022openvocabulary} except for a shorter schedule (due to frozen backbone) and a background weight following~\citet{Zareian_2021_CVPR,zhong2021regionclip} (see~\ref{sec:background-loss}). The hyper-parameter differences on COCO are to mitigate overfitting to the ZSD-COCO split~\citep{bansal2018zero} of 48 categories, which is significantly smaller than the 800 LVIS base categories. This is necessary because F-VLM does not use other objectives \eg knowledge distillation or weak supervision to counter-balance overfitting.
\begin{table}[h]
\caption{\textbf{\OURS{} hyper-parameter configuration.}}
\centering
\begin{tabular}{lcc}
\textbf{Configuration}  & \textbf{LVIS} & \textbf{COCO} \\
\toprule
Optimizer & SGD & SGD \\
Momentum & $\beta=0.9$ & $\beta=0.9$ \\
Weight decay & 1e-4 & 1e-2 \\
Gradient Clipping & none & none \\
Learning rate (LR) & 0.36 & 0.02 \\
Step decay factor & 0.1$\times$ & 0.1$\times$ \\
Step decay schedule & [0.8, 0.9, 0.95] & [0.9, 0.95, 0.975] \\
Warmup LR / steps & 3.2e-3 / 1k &  3.2e-3 / 1k  \\
Total Steps & 46.1k & 11.25k \\
Batch size & 256 & 64 \\
Epochs & 118 & 6 \\
Augmentation & LSJ~\citep{ghiasi2021simple} & LSJ~\citep{ghiasi2021simple} \\
NMS Threshold & 0.5 & 0.4\\
Base VLM weight $\alpha$ & 0.35 & 0.2 \\
Novel VLM weight $\beta$ & 0.65 & 0.45\\
Background weight $\gamma$ & 0.9 & 0.2 \\
VLM Temperature $T$& 0.01 & 0.01 \\
\bottomrule
\end{tabular}
\label{tab:hyperparams}
\end{table}

We observe some differences in the optimal hyper-parameters for different backbone architectures. With the R50x64 backbone, we notice an improvement of 1.0 AP$_r$ when we use $T=0.02$ as opposed to the default $T=0.01$. For R50 backbone, we notice an improvement of 0.5 AP$_r$ when we apply a gradient clipping of 1.0 maximum gradient norm as opposed to none. We report the performance using the optimal settings for these two backbones.

\section{Visualization}
\label{sec:more-vis}
\begin{figure*}[t]
    \centering
    \begin{subfigure}[b]{0.95\linewidth}
        \centering
    	\includegraphics[width=0.98\linewidth]{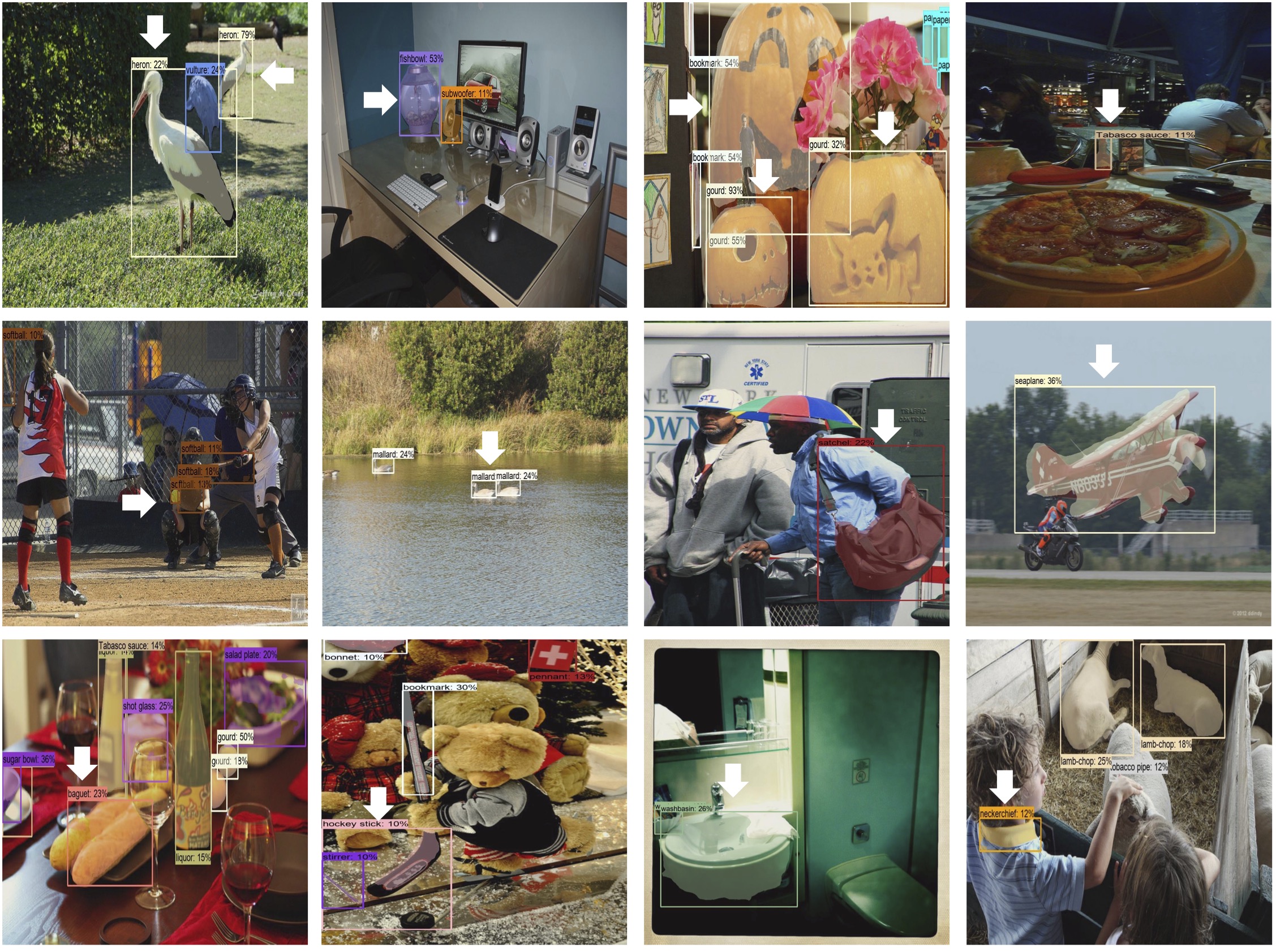}
    	\caption{\textbf{LVIS novel category detection}. \OURS{} can detect many novel categories despite its simplicity using a frozen VLM. The white arrows point to the novel objects correctly detected by \OURS{}. For clarity, we only show the novel categories.}
    	\label{fig:supp-lvis}
    \end{subfigure}
    \begin{subfigure}[b]{0.95\linewidth}
        \centering
    	\includegraphics[width=0.98\linewidth]{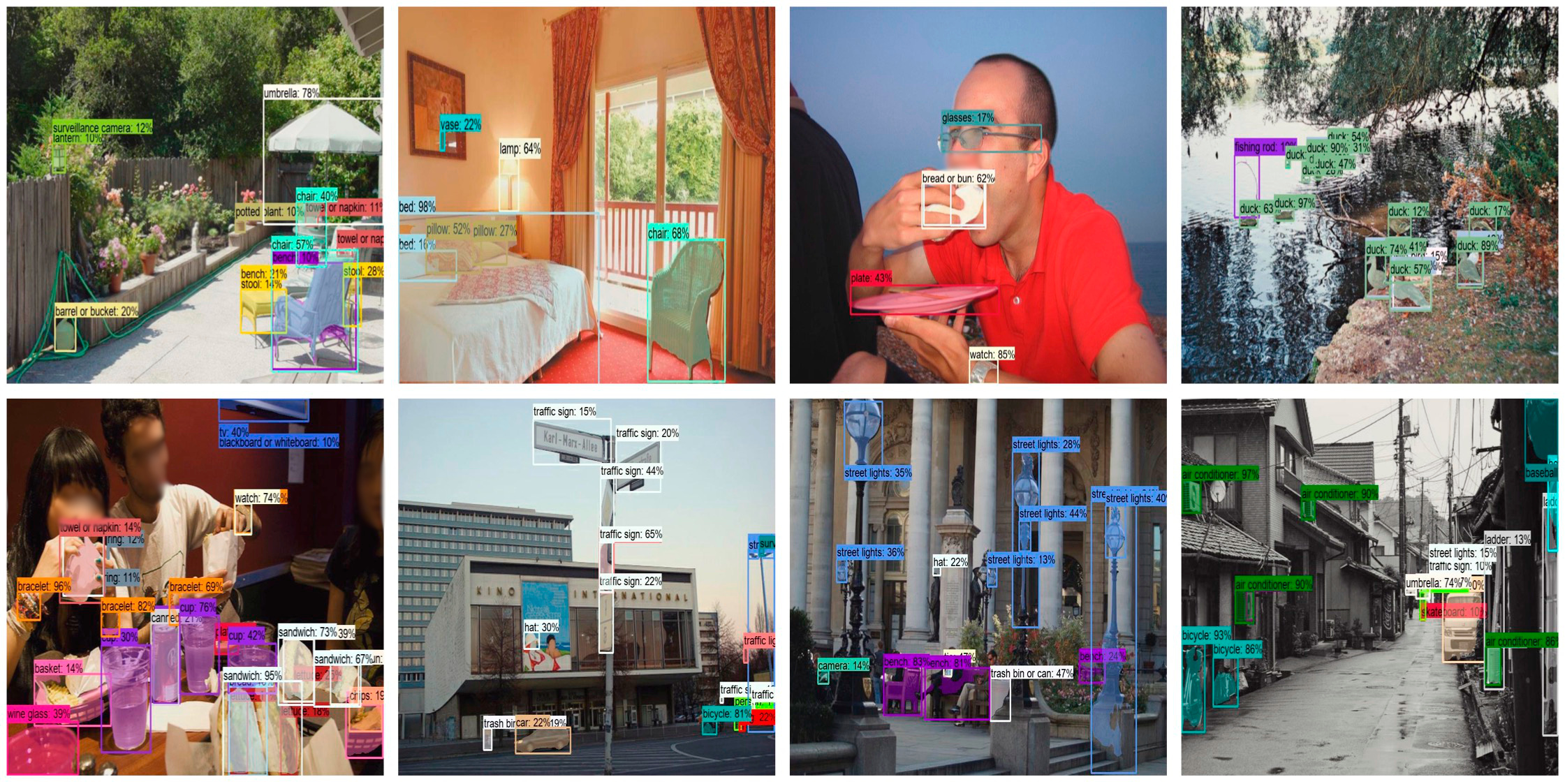}
    	\caption{\textbf{Objects365 transfer object detection}. \OURS{} can be applied to a new dataset and detect many challenging categories without further finetuning.}
    	\label{fig:supp-o365}
	\end{subfigure}
	\caption{\OURS{} Visualization on LVIS novel categories and Objects365 transfer object detection.}
	\label{fig:fvlm-supp-vis}
\end{figure*}

We visualize more \OURS{} outputs on LVIS novel categories and transfer detection to Objects365 in Figure~\ref{fig:fvlm-supp-vis}. On LVIS, \OURS{} is able to correctly detect many rare categories including baguet, neckerchief, tabasco sauce, and gourd. On Objects365, \OURS{} can detect many categories in complex scenes including ducks, traffic sign, street light, and air conditioner. These confirm that our approach is effective for novel category detection and transfer detection to another dataset. We use the R50x4 backbone for this visualization. The model was trained on the LVIS base categories following the main benchmark of the paper. 

\section{Application on Ego-Centric Data} 
\label{sec:ego4d}

A key benefit of open-vocabulary detection is to test on out-of-distribution data with categories given by users on the fly. Thus, we apply \OURS{} to Ego4D~\citep{Grauman_2022_CVPR}, a real-world ego-centric application. We train \OURS{} on a mixture of full LVIS, Objects365, and COCO datasets to expand its training vocabulary for application in the wild, and use the R50x16 backbone for this experiment. The model is not trained on Ego4D in order to evaluate for transfer detection. The categories are provided by the user based on visual inspection of the video. 

For the indoor scene, the category names provided by the user are as follows: \textit{plate, cabinet, stove, towel, cleaning rag, ventilator, knob, sauce and seasoning, steel lid, window, window blinds, plant, light switch, light, door, carpet, exit sign, doormat, hair, door lock, tree, poster on the wall, sticker on the wall, faucet, recycle bin, rack, hand, can, carton, trash, Christmas tree, plastic container, fridge}.

For the grocery store scene, the category names provided by the user are as follows: \textit{exit sign, poster, chocolate bar, bag of candy, bag of cookies, snack, oreo, soy sauce, apple, pear, orange, grapes, price tag, cereal, instant noodle/ramen, cracker, ATM  machine, instant noodle, wooden basket, red ramen bowls, magazine, drugs and medicine, Mayo, Ketchup, Cup noodle, burrito, Lays/Sun chips, seasoning sauce, black carton, salad dressing, canned food}. 

\begin{figure*}[t]
    \centering
    \begin{subfigure}[b]{0.95\linewidth}
        \centering
    	\includegraphics[width=0.98\linewidth]{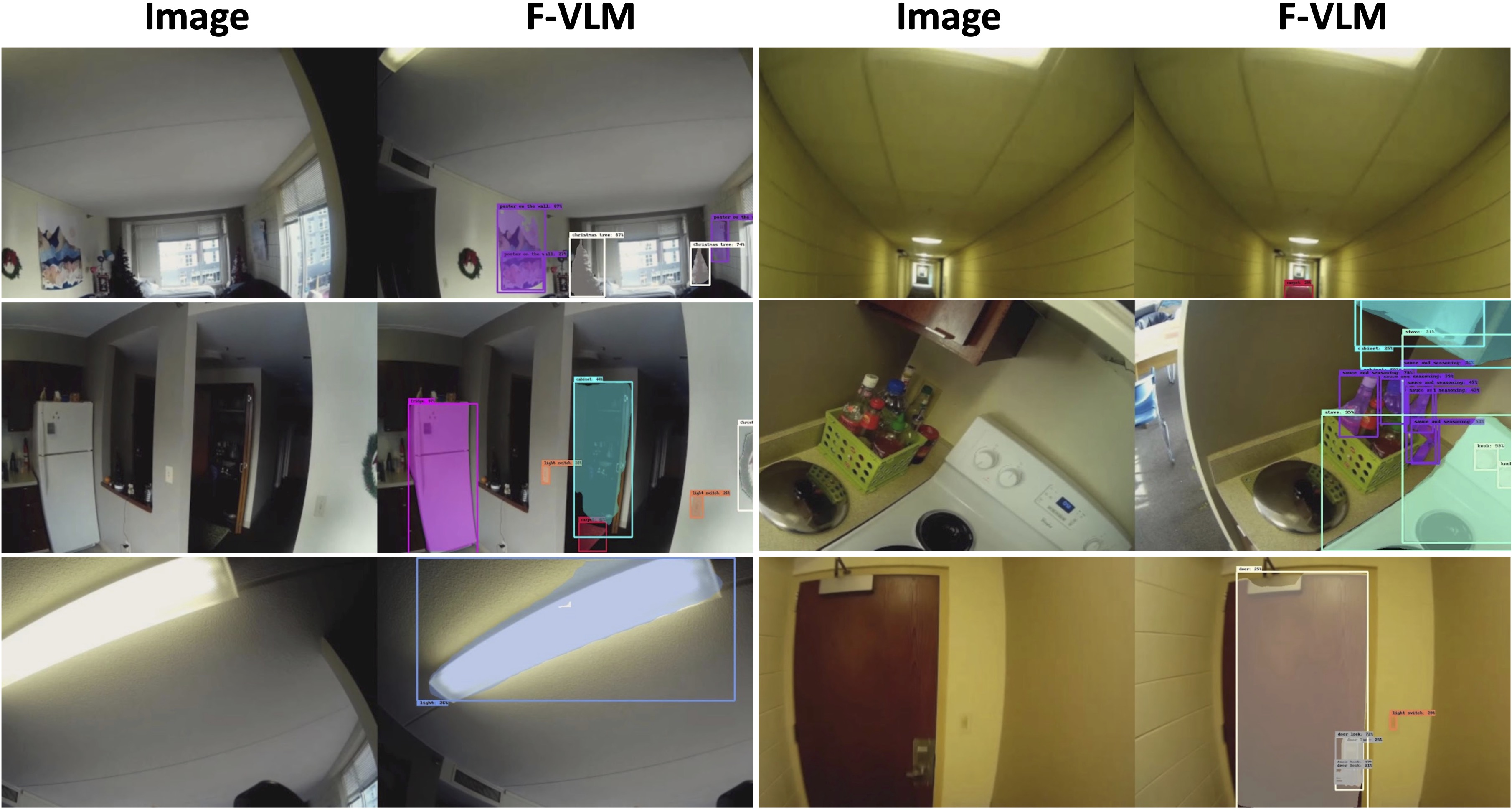}
    	\caption{\textbf{Indoor scene.} Under the challenging viewing angles, occlusion, and lighting conditions, \OURS{} still manages to detect many objects in the scene.}
    	\vspace{4mm}
    	\label{fig:ego-indoor}
    \end{subfigure}
    \begin{subfigure}[b]{0.95\linewidth}
        \centering
    	\includegraphics[width=0.98\linewidth]{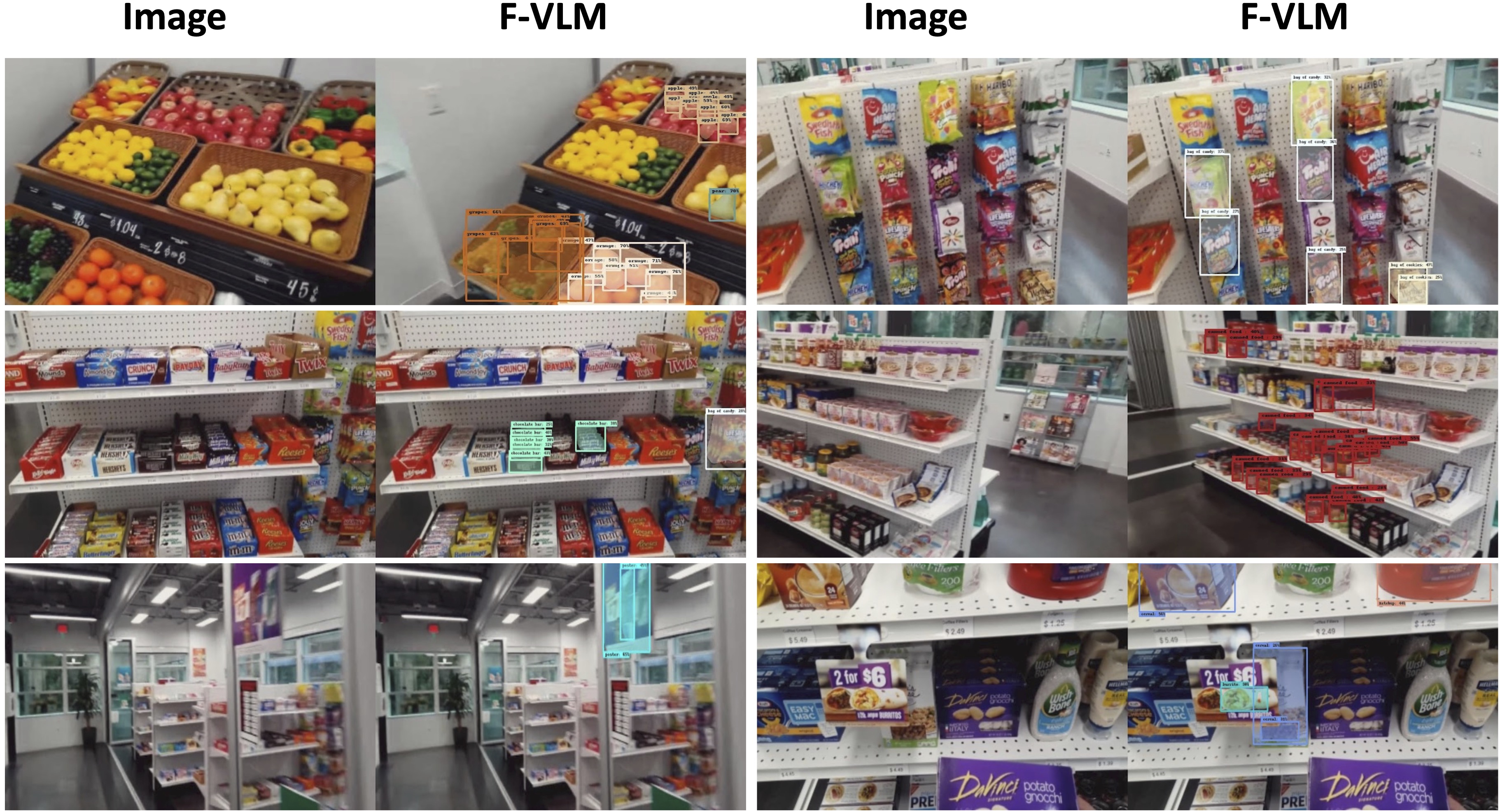}
    	\caption{\textbf{Grocery store scene}. The scene is very crowded with a wide variety of objects. \OURS{} is able to detect many of them.}
    	\label{fig:ego-grocery}
	\end{subfigure}
	\caption{\OURS{} Visualization on Ego4D~\citep{Grauman_2022_CVPR} transfer object detection. Novel categories detected: \textit{light switch, light, door lock, sauce and seasoning, bag of candies, canned food, and burrito}.}
	\label{fig:ego4d-supp-vis}
\end{figure*}

Figure~\ref{fig:ego4d-supp-vis} shows that \OURS{} is able to detect many objects in the ego-centric videos despite the large domain shift and challenging viewing conditions. In particular, it is able to detect many novel categories not present in the training set, such as light switch, light, door lock, sauce and seasoning, bag of candies, canned food, and burrito.

\end{document}